\documentclass[letterpaper, 10 pt, journal, twoside]{ieeetran}

\usepackage{amsmath,amsfonts}
\usepackage{algorithmic}
\usepackage{array}
\usepackage{textcomp}
\usepackage{stfloats}
\usepackage{url}
\usepackage{verbatim}
\usepackage{graphicx}
\def\BibTeX{{\rm B\kern-.05em{\sc i\kern-.025em b}\kern-.08em
    T\kern-.1667em\lower.7ex\hbox{E}\kern-.125emX}}
\usepackage{balance}

\usepackage{amsfonts}
\usepackage{amssymb}
\usepackage{mathtools}
\usepackage{bm}  
\usepackage{graphics} 
\usepackage{epsfig} 
\usepackage{wrapfig}
\usepackage{subcaption}
\usepackage[font=small,labelfont=small]{caption}
\usepackage{color}
\usepackage{algorithm}
\usepackage{algorithmic}
\usepackage{sidecap}
\sidecaptionvpos{figure}{c}
\usepackage{tikz}
\usetikzlibrary{positioning}
\usepackage{tabularx}
\usepackage{colortbl}
\usepackage{booktabs}
\usepackage{multirow}
\usepackage{siunitx}
\usepackage{nicefrac}
\usepackage[nolist]{acronym} 
\usepackage{cuted}
\usepackage[noadjust]{cite}  

\usepackage{amsthm} 
\usepackage[framemethod=tikz]{mdframed} 
\usepackage{lipsum}

\usepackage{soul} 
\usepackage[dvipsnames]{xcolor}
\usepackage[normalem]{ulem}

\usepackage{amsmath}

\newcommand{\trsp}{\mathsf{T}}  

\newcommand{\euclideanspace}{\mathbb{R}}
\newcommand{\manifold}{\mathcal{M}}
\newcommand{\tangentspace}[1]{\mathcal{T}_{#1}\mathcal{M}}
\newcommand{\tangentbundle}{\mathcal{TM}}

\newcommand{\innerprod}[3]{\langle #2, #3 \rangle_{#1}}  

\newcommand{\specialorth}[1]{\operatorname{SO}(#1)}
\newcommand{\specialeucl}[1]{\operatorname{SE}(#1)}

\newcommand{\configmanifold}{\mathcal{Q}}
\newcommand{\configtangentbundle}{\mathcal{T}\configmanifold}

\newcommand{\configtangentspace}[1]{\mathcal{T}_{#1}\configmanifold}

\newcommand{\taskmanifold}{\mathcal{X}}
\newcommand{\tasktangentspace}[1]{\mathcal{T}_{#1}\taskmanifold}
\newcommand{\tasktangentbundle}{\mathcal{T}\taskmanifold}

\newcommand{\group}{\mathbb{G}}
\newcommand{\groupon}[1]{\group^{\scriptscriptstyle #1}}
\newcommand{\groupelementof}[1]{g_{\scriptscriptstyle #1}}
\newcommand{\groupactionblank}{\Phi}
\newcommand{\groupactiononblank}[1]{\Phi^{\scriptscriptstyle #1}}
\newcommand{\groupactionon}[2]{\Phi^{\scriptscriptstyle #1}_{#2}}
\newcommand{\groupaction}[1]{\Phi_{#1}}

\newcommand{\vecfield}[1]{X^{\scriptscriptstyle #1}}
\newcommand{\infgenerator}[2]{X^{\scriptscriptstyle #1}_{#2}}

\newcommand{\orthogonalgroup}[1]{\text{O}(#1)}
\newcommand{\specorthogonalgroup}[1]{\text{SO}(#1)}
\newcommand{\specorthogonaliealgebra}[1]{\mathfrak{so}(#1)}
\newcommand{\cyclicgroup}[1]{\mathbb{C}_{#1}}
\newcommand{\scalinggroup}[1]{\mathbb{S}_{#1}}

\newcommand{\liealgebrablank}{\mathfrak{g}}

\newcommand{\vertangent}[1]{\mathcal{V}_{#1}}
\newcommand{\hortangent}[1]{\mathcal{H}_{#1}}

\newtheorem{example}{Example}[section]
\theoremstyle{definition}
\newtheorem{definition}{Definition}[section]

\newtheorem{proposition}{Proposition}[section]

\usepackage[colorlinks=True,
            linkcolor=blue,
            citecolor=blue,
            urlcolor=blue]{hyperref}

\definecolor{lightlightgray}{rgb}{0.92, 0.92, 0.92}
\definecolor{lightsteelblue}{rgb}{0.7 0.77 0.87}
\definecolor{lightlightsteelblue}{rgb}{0.88 0.92 0.97}
\definecolor{colDemos}{HTML}{444444}
\definecolor{colBaseline}{HTML}{E09F1B}
\definecolor{colSO2}{HTML}{17B2FF}
\definecolor{colSO2S2}{HTML}{7BB36A}
\definecolor{colFull}{HTML}{BC34D1}

\definecolor{colSO5Deg}{HTML}{2196F3}
\definecolor{colSO10Deg}{HTML}{E63946}
\definecolor{colSO15Deg}{HTML}{FF9800}
\definecolor{colSO30Deg}{HTML}{E91E63}
\definecolor{colSO45Deg}{HTML}{2CA02C}
\definecolor{colSO60Deg}{HTML}{8B4513}
\definecolor{colSO75Deg}{HTML}{00BCD4}
\definecolor{colSO90Deg}{HTML}{9C27B0}

\definecolor{bestcell}{HTML}{F0F7EE}  

\DeclareRobustCommand{\blackline}{\raisebox{2pt}{\tikz{\draw[black,solid,line width = 1.5pt](0,0) -- (3mm,0);}}}
\DeclareRobustCommand{\grayline}{\raisebox{2pt}{\tikz{\draw[gray,solid,line width = 1.5pt](0,0) -- (3mm,0);}}}
\DeclareRobustCommand{\baselineline}{\raisebox{2pt}{\tikz{\draw[colBaseline,solid,line width = 1.5pt](0,0) -- (3mm,0);}}}
\DeclareRobustCommand{\SOline}{\raisebox{2pt}{\tikz{\draw[colSO2,solid,line width = 1.5pt](0,0) -- (3mm,0);}}}
\DeclareRobustCommand{\SOSline}{\raisebox{2pt}{\tikz{\draw[colSO2S2,solid,line width = 1.5pt](0,0) -- (3mm,0);}}}
\DeclareRobustCommand{\allsymline}{\raisebox{2pt}{\tikz{\draw[colFull,solid,line width = 1.5pt](0,0) -- (3mm,0);}}}

\DeclareRobustCommand{\SOFiveDeg}{\raisebox{2pt}{\tikz{\draw[colSO5Deg,solid,line width = 1.5pt](0,0) -- (3mm,0);}}}
\DeclareRobustCommand{\SOTenDeg}{\raisebox{2pt}{\tikz{\draw[colSO10Deg,solid,line width = 1.5pt](0,0) -- (3mm,0);}}}
\DeclareRobustCommand{\SOFifteenDeg}{\raisebox{2pt}{\tikz{\draw[colSO15Deg,solid,line width = 1.5pt](0,0) -- (3mm,0);}}}
\DeclareRobustCommand{\SOThirtyDeg}{\raisebox{2pt}{\tikz{\draw[colSO30Deg,solid,line width = 1.5pt](0,0) -- (3mm,0);}}}
\DeclareRobustCommand{\SOFortyFiveDeg}{\raisebox{2pt}{\tikz{\draw[colSO45Deg,solid,line width = 1.5pt](0,0) -- (3mm,0);}}}
\DeclareRobustCommand{\SOSixtyDeg}{\raisebox{2pt}{\tikz{\draw[colSO60Deg,solid,line width = 1.5pt](0,0) -- (3mm,0);}}}
\DeclareRobustCommand{\SOSeventyFiveDeg}{\raisebox{2pt}{\tikz{\draw[colSO75Deg,solid,line width = 1.5pt](0,0) -- (3mm,0);}}}
\DeclareRobustCommand{\SONinetyDeg}{\raisebox{2pt}{\tikz{\draw[colSO90Deg,solid,line width = 1.5pt](0,0) -- (3mm,0);}}}

\definecolor{rot_vec_field_color}{HTML}{1f5fa6}
\definecolor{scaling_vec_field_color}{HTML}{d1495b}

\begin{document}

\title{Symmetries Here and There, Combined Everywhere: \\Cross-space Symmetry Compositions in Robotics}

\author{Loizos Hadjiloizou, Rodrigo P\'{e}rez-Dattari, and No\'{e}mie Jaquier%
\thanks{This work was supported by the Wallenberg Artificial Intelligence, Autonomous Systems, and Software Program funded by the Knut and Alice Wallenberg Foundation.} 
\thanks{All authors are with the Department of Robotics, Perception and Learning, KTH Royal Institute of Technology.
        {\tt\footnotesize \{\href{mailto:loizosh@kth.se}{\textrm{loizosh}}, \href{mailto:rpd@kth.se}{\textrm{rpd}}, \href{mailto:jaquier@kth.se}{\textrm{jaquier}}\}@kth.se }}%
}

\maketitle
\begin{abstract}
Robots exhibit a rich variety of symmetries arising from their mechanical structure and the properties of their tasks.
Although many robotics problems exhibit several symmetries simultaneously, existing approaches typically treat them in isolation, failing to exploit their combined potential. This paper introduces cross-space symmetry compositions, a framework for learning robot policies that are jointly equivariant to multiple symmetries across configuration and task spaces. Leveraging the differential-geometric structure of the forward kinematics map, we both descend symmetries from configuration to task space and lift symmetries from task to configuration space, enabling their composition within a unified representation space. We validate our framework on simulated and real-world experiments on a dual-arm robot, demonstrating that jointly leveraging multiple symmetries yields improved generalization.
Video and source code are available at \url{https://sites.google.com/view/cross-space-symmetries}.
\end{abstract}

\section{Introduction}
\label{sec:intro}

\IEEEPARstart{I}{ncorporating} inductive bias is a key avenue to improve sample efficiency and generalization in robot policy learning. Among possible biases, those arising from \emph{symmetries} are particularly powerful as they enable a policy to generalize from each training sample to an entire class of equivalent observations and actions~\cite{otto2025unified}.  
In robot policy learning, a symmetry describes a transformation of the observation inducing an associated transformation of the action, leading to consistent behavior across symmetric situations. Symmetries may arise from physical laws and mechanical structures~\cite{marsden2016introduction} or may be identified directly from data as invariance and equivalence relations that a policy should respect~\cite{desai2022symmetry}.

Robots exhibit a rich variety of symmetries rooted both in their mechanical structure and in the properties of their tasks. These symmetries have been largely used in robot control and learning. In configuration space, the Lagrangian of floating-based robots is invariant to translations and reorientations of their base~\cite{marsden2016introduction}. Factoring out these transformations from the equations of motion enables control and learning on lower-dimensional spaces~\cite{ostrowski1999computing,ohsawa2013symmetry, mishra2022reduced, welde2025_RLsymmetries}. Morphological symmetries~\cite{apraez2025morphological} arise from the interchangeability of replicated kinematic chains such a humanoid robot's arms or a quadruped's legs. They were exploited via data augmentation procedures and equivariant neural architectures for state estimation~\cite{apraez2025morphological} and policy learning~\cite{li2025morphologically}, yielding enhanced sample efficiency and stronger generalization than symmetry-unaware algorithms. 
Symmetries in configuration space were also shown to improve consistency and reduce sample complexity in motion planning~\cite{cohn2025sampling-based}.
In task space, symmetries capture equivalences in task specifications. Euclidean isometries, e.g., translations and rotations, are widely used to learn equivariant grasping and manipulation policies that generalize to unseen object poses and workspace configurations~\cite{zeng2021transporter,zhu2022:GraspEquivariant,wang2022:so2equivariant,ryu2023equivariant,huang2024leveraging}. Many robotics tasks exhibit task-induced symmetries~\cite{kallem2010task}, where states or actions achieving the same objective are considered equivalent.\looseness-1

\begin{figure}[t]
    \centering
    \includegraphics[width=\linewidth, trim=6.65cm 5.15cm 6.65cm 5.33cm, clip]{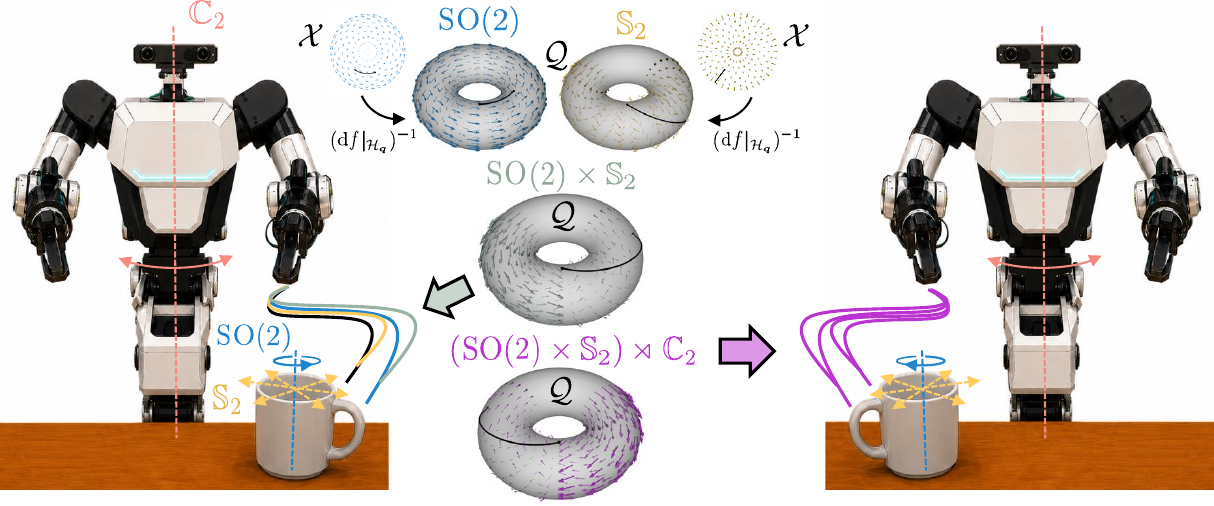}
    \caption{Manipulation task with several symmetries. We lift the rotation and scaling symmetries from task space $\taskmanifold$ to configuration space $\configmanifold$, where we compose them with the morphological symmetry. The learned policy allows the robot to pick the mug (\blackline) under symmetric transformations via scaling (\baselineline), rotation (\SOline), scaling and rotation (\SOSline), and their compositions with morphological reflection (\allsymline).  }
    \label{fig:symmetries}
\vspace{-0.7cm}
\end{figure}

In many manipulation tasks, robots exhibit several of the aforementioned symmetries simultaneously (see Fig.~\ref{fig:symmetries}). Yet, previous works consider them in isolation, overlooking relevant inductive bias. For instance, a dual-arm policy encoding rotational symmetry generalizes to rotated scenes but not to a reflection of the workspace. Conversely, a morphologically-symmetric policy only exploits limbs interchangeability, but fails under scene rotations. Robust generalization across both variations thus requires jointly encoding both symmetries. Compositions of symmetries have been explored in machine learning. Kim \emph{et~al.}~\cite{kim23:mixed-symmetries} weakly enforced multiple symmetries through regularizers in the objective function. In a similar line, Li \emph{et~al.}~\cite{li2025:latent-mixture-symmetries} learned dynamic models via latent flows that preserve several symmetries simultaneously. However, both considered symmetries arising in a single space. In contrast, robotics symmetries naturally emerge in different spaces: Floating-base and morphological symmetries are defined in configuration space, while task isometries and task-induced symmetries arise in task space. Thus, they cannot be composed directly and must first be transferred in a common space.

In this paper, we propose \emph{cross-space symmetry compositions}, a framework for learning robot policies that are jointly equivariant to multiple symmetries across configuration and task spaces (Sec.~\ref{sec:catalog_of_symmetries}). Our approach builds on two key components: \emph{(1)} The transfer of symmetries into the configuration or the task space (Sec.~\ref{sec:transferring_symmetries}); and \emph{(2)} Their composition within that space (Sec.~\ref{sec:symmetry_composition}).
Taking a vector-field view on symmetries and leveraging the differential-geometric structure of forward kinematics, we propose a unified framework to \emph{descend} symmetries from configuration to task space and \emph{lift} symmetries from task to configuration space. We show that descending a configuration-space symmetry reduces to verifying the equivariance of the forward kinematics map, and lift task-space symmetries by identifying the aforementioned map as a \emph{smooth submersion}.  
We characterize the conditions under which the transferred symmetries are systematically composed within a common space through direct or semi-direct products. 
We validate our approach via simulated and real-world experiments on a dual-arm manipulator, showing that jointly leveraging multiple symmetries improves generalizability (Sec.~\ref{sec:experimental_validation}).\looseness-1

\vspace{-0.15cm}
\begin{mdframed}[hidealllines=true,backgroundcolor=lightlightsteelblue,innerleftmargin=.1cm,innerrightmargin=.1cm,innertopmargin=.1cm,innerbottommargin=0.1cm,, roundcorner=2pt]
In summary, we contribute: \emph{(1)} A unified review of the principal families of symmetries arising in robotics, spanning both configuration and task spaces; \emph{(2)} A cross-space symmetry transfer framework to descend and lift symmetries between configuration and task spaces, together with feasibility conditions; \emph{(3)} A framework for composing multiple symmetries within a common space, including a characterization of the necessary compatibility conditions.
\end{mdframed}
\vspace{-0.1cm}

\vspace{-0.2cm}
\section{Preliminaries}
\vspace{-0.05cm}
\label{sec:preliminaries}

\subsection{Differential Geometry}
Intuitively, a smooth manifold $\manifold$ of dimension $n$ is a topological space that locally resembles the Euclidean space $\mathbb{R}^{n}$. Formally, there exists a neighborhood homeomorphic to an open subset of $\mathbb{R}^{n}$ around each point $\bm{p} \!\in\! \manifold$. 
The tangent space $\tangentspace{\bm{p}}$ at every point $\bm{p} \!\in\! \manifold$ is an $n-$dimensional vector space composed of the first-order derivatives of curves on the manifold passing through $\bm{p}$.
The Euclidean structure of tangent spaces enables the use of tools from calculus, including differentiation, integration, and vector fields.
The tangent bundle $\tangentbundle$ is the disjoint union of all tangent spaces. \looseness-1

\subsection{Groups, Actions, and Equivariance}
\label{subsec:groups}

Symmetries are naturally represented through the mathematical structure of \emph{groups} and their associated \emph{group actions}.
A group $\group$ is a set equipped with an associative binary operation $\circ \!:\! \group \! \times \! \group \! \to \! \group$, an identity element $e\in \group$, and an inverse $g^{-1} \in \group$ for each element $g\in\group$~\cite{hall2015lie}. A group acts on a manifold $\manifold$ via its action $\Phi$, which defines transformations parametrized by the group elements. We focus on the \emph{left action}, which is an associative map $\groupactiononblank{\manifold} \!:\!\group \!\times\! \manifold \!\to\! \manifold$ that respects the identity transformation $\groupactiononblank{\manifold}(e,\bm{p})\!=\!\bm{p}$, $\forall \bm{p}\!\in\!\manifold$.
A group action is \emph{global} if it is defined for every $(g, \bm{q}) \in \group \!\times\! \configmanifold$~\cite[Def.~1.25]{olver1993applications}.
Intuitively, a map ${h\!:\!\manifold \!\to\! \mathcal{N}}$ across two manifolds $\manifold$, $\mathcal{N}$ is
$\group$-equivariant when transforming the input before applying the map is equivalent to first applying the map and subsequently transforming the output. 
Mathematically, $h$ is \emph{$\group$-equivariant} if\looseness-1
\begin{equation}
    h\big(\groupactiononblank{\manifold}(g, \bm{p}) \big) = \groupactiononblank{\mathcal{N}}\big(g, h(\bm{p}) \big) 
    \qquad 
    \forall g \in \group \; \bm{p}\in\manifold.
    \label{eq:equivariance_condition}
\end{equation}
The \emph{equivariance condition}~\eqref{eq:equivariance_condition} underpins equivariant learning models, where it is used for data augmentation~\cite{luo2024reinforcement}, to penalize symmetry violations via regularization~\cite{wang2022approximately}, or to enforce a symmetry directly in the network architecture~\cite{finzi2021practical}.

\textbf{Finite and Lie Groups}. Robotic systems exhibit both \emph{discrete} and \emph{continuous} symmetries. Discrete symmetries such as morphological symmetries are described by \emph{finite groups}, i.e., groups containing finitely-many elements. Continuous symmetries, such as translational and rotational symmetries, are described by \emph{Lie groups}. A Lie group $\group$ is simultaneously a group and a smooth manifold such that the group operations are smooth.
Its tangent space at the identity $\liealgebrablank = \mathcal{T}_{\bm{e}}\group$, known as the \emph{Lie algebra}, provides an infinitesimal description of $\group$.
The Lie group and its Lie algebra are related through the exponential map $\operatorname{exp} \!: \!\liealgebrablank\! \rightarrow \!\group$.
A Lie group action on $\manifold$ induces actions on its tangent spaces, transforming velocities via its differential $\mathrm{d}\Phi^{\scriptscriptstyle \manifold}_{g}|_{\bm{p}} \!:\! \tangentspace{\bm{p}} \!\to\! \tangentspace{\groupactiononblank{\manifold}(g, \bm{p})}$ with $g \!\in\! \group$, $\bm{p} \!\in\! \manifold$. For instance, the special orthogonal group $\specorthogonalgroup{2}$ acting on Euclidean space rotates both positions and the associated velocities.

\textbf{Infinitesimal description of group actions.} Continuous symmetries can also be characterized from an infinitesimal perspective via vector fields on $\manifold$. In particular, each Lie algebra element  $\bm{\xi}\in\liealgebrablank$ induces a vector field ${\infgenerator{\manifold}{\bm{\xi}}\!:\!\manifold\!\to\!\tangentbundle}$, known as its \emph{infinitesimal generator}, given at $\bm{p}\in\manifold$ as
\begin{equation}
    \label{eq:infinitesimal_generator_from_group_action}
    \infgenerator{\manifold}{\bm{\xi}}(\bm{p}) = \frac{d}{dt}\Big|_{t=0} \groupactiononblank{\manifold}(\exp(t\bm{\xi}), \bm{p}).
\end{equation}
The flow generated by $\infgenerator{\manifold}{\bm{\xi}}$ corresponds to applying the actions $\groupactiononblank{\manifold}(\exp(t\bm{\xi}), \bm{p})$. For instance, the action of the special orthogonal group $\specorthogonalgroup{2}$ on $\mathbb{R}^{2}$ induces rotational vector fields tangent to circles around the origin. 

\textbf{Lie bracket.}
Two smooth vector fields $X^{\scriptscriptstyle \manifold}$, $Y^{\scriptscriptstyle \manifold}$ on a manifold $\manifold$ can be combined into another vector field via their \emph{Lie bracket}. In local coordinates, the Lie bracket is 
\begin{equation}
    [X^{\scriptscriptstyle \manifold}, Y^{\scriptscriptstyle \manifold}](\bm{p}) =
    \mathrm{d}X^{\scriptscriptstyle \manifold}|_{\bm{p}}Y^{\scriptscriptstyle \manifold}(\bm{p}) - 
    \mathrm{d}Y^{\scriptscriptstyle \manifold}|_{\bm{p}}X^{\scriptscriptstyle \manifold}(\bm{p}),
\end{equation}
for $\bm{p} \in \manifold$. 
It measures the non-commutativity of the flows generated by the vector fields, i.e., they commute if and only if $[X^{\scriptscriptstyle \manifold}, Y^{\scriptscriptstyle \manifold}](\bm{p}) \!=\!0, \forall \bm{p} \in \manifold$. 

\subsection{Riemannian Manifolds and Maps}
\label{subsec:RiemannianBackground}
A Riemannian manifold $(\manifold, m)$ is a smooth manifold $\manifold$ equipped with a smoothly-varying inner product $m$ called a Riemannian metric and given as ${\innerprod{\manifold}{\dot{\bm{p}}_1}{\dot{\bm{p}}_2}\!=\!\dot{\bm{p}}_1^\trsp \bm{M}(\bm{p})\dot{\bm{p}}_2}$,  $\dot{\bm{p}}_1,\dot{\bm{p}}_2\!\in\!\tangentspace{\bm{p}}$ in coordinates.
A smooth map $h\!:\!\manifold \!\to\! \mathcal{N}$ between two Riemannian manifolds is a \emph{smooth submersion} at $\bm{p}\!\in\!\manifold$ if its differential ${\mathrm{d} h|_{\bm{p}} \!:\! \tangentspace{\bm{p}} \!\rightarrow\! \mathcal{T}_{h(\bm{p})}\mathcal{N}}$ is surjective. A smooth submersion decomposes each tangent space as $\tangentspace{\bm{p}} \!=\! \mathcal{V}_{\bm{p}} \!+\! \mathcal{H}_{\bm{p}}$~\cite{lee2012introduction}, with vertical subspace ${\mathcal{V}_{\bm{p}} = \operatorname{Ker}(\mathrm{d}h)|_{\bm{p}} = \{\bm{v} \in \tangentspace{\bm{q}} : \mathrm{d}h|_{\bm{p}}(\bm{v})=0\}}$. The horizontal subspace $\mathcal{H}_{\bm{p}} = \mathcal{V}_{\bm{p}}^{\bot}$ is its $m$-orthogonal complement.\looseness-1

The robot configuration space $(\configmanifold, m_{\configmanifold})$ and task space $(\taskmanifold, m_{\taskmanifold})$ are Riemannian manifolds whose elements are joint configurations and end-effector poses, respectively. 
They are related through the forward kinematics map ${f \!:\! \configmanifold \!\rightarrow\! \taskmanifold}$
In coordinates, its differential ${\mathrm{d} f|_{\bm{q}} \!:\! \configtangentspace{\bm{q}} \!\rightarrow\! \mathcal{T}_{f(\bm{q})}\taskmanifold}$ is the robot Jacobian $\bm{J}(\bm{q})$.
In this paper, we utilize that, under some conditions, the forward kinematics map of redundant robots is a smooth submersion to lift symmetries from $\taskmanifold$ to $\configmanifold$. 

\section{A Catalog of Robotic Symmetries}
\label{sec:catalog_of_symmetries}

We review the principal families of symmetries commonly encountered in robotics, summarized in Table~\ref{tab:symmetries}. We categorize configuration-space and task-space symmetries which act on the robot configuration and task space, respectively.

\begin{table}[t]
    \centering
    \caption{Summary of principal families of robotic symmetries.}
    \label{tab:symmetries}
    \setlength{\tabcolsep}{3pt}
    \resizebox{\linewidth}{!}{
    \begin{tabular}{lccccc}
        \toprule
        \textbf{Symmetry}
            & \textbf{Origin}
            & \textbf{Acts on} 
            & \textbf{Preserves}
            & \textbf{Transformation}
            &  \\
        \midrule
        Morphological & Mec. struct. & $\configmanifold$ & Lagrangian $\mathcal{L}$ & Group $\group_{\text{M}}$ & \\
        Floating-based & Space geometry & $\configmanifold$ & Lagrangian $\mathcal{L}$ & Group $\group_{\text{F}}$ & \\
        \midrule
        Rotational & Space geometry & $\taskmanifold$ & Eucl. metric, origin & Group $\group_{\text{R}}=\specorthogonalgroup{n}$ & \\
        Task-induced & Task spec. & $\taskmanifold$ & Cost $c$ & Vec. field $\vecfield{\taskmanifold}$ (group $\group_{\text{T}}$) & \\
        \bottomrule
    \end{tabular}
    }
    \vspace{-0.5cm}
\end{table}

\subsection{Morphological and Floating-based Symmetries}

Morphological symmetries~\cite{apraez2025morphological} capture the interchangeability of replicated kinematic chains with symmetric mass distributions, such as the arms of a humanoid robot or the legs of a quadruped. These symmetries arise from the robot's mechanical structure, preserve the robot's Lagrangian, and induce equivariances in both the state space and dynamics equations.
Morphological symmetries naturally act on the robot configuration space $\configmanifold$.
Morphological symmetries of a rigid-bodied robot with fixed morphology (i.e., with finitely-many interchangeable kinematic chains) are characterized as a finite group $\group_{\text{M}}$ acting on $\configmanifold$ through the group action
\begin{equation}
    \label{eq:morphosymm:config_action}
    \groupactionon{\configmanifold}{\text{M}}: \group_{\text{M}} \times \configmanifold \rightarrow \configmanifold, \quad\quad \groupactionon{\configmanifold}{\text{M}}(\groupelementof{\text{M}}, \bm{q}) = \bm{\rho}_{\configmanifold}(\groupelementof{\text{M}})\bm{q},     
\end{equation}
where the matrix $\bm{\rho}_{\configmanifold}(\groupelementof{\text{M}})$ is built from correspondences between interchangeable kinematic chains and defines a non-trivial transformation of the joint coordinates. 
For robots whose symmetric chains are composed of $1$-dimensional revolute and prismatic joints, $\bm{\rho}_{\configmanifold}$ is a signed permutation matrix. It exchanges each joint coordinate with its morphologically-symmetric counterpart, applying a sign inversion whenever paired joints have opposite coordinate conventions. For example, it swaps the joint coordinates of the two arms of a humanoid robot to mimic a reflection across the sagittal plane.\looseness-1

For floating-based robots, morphological symmetries are often paired with floating-based symmetries~\cite{marsden2016introduction} to act via\looseness-1
\begin{equation}
    \label{eq:morphosymm:base-config_action}
    \groupactionon{\configmanifold}{\text{M}}(\groupelementof{\text{M}}, \bm{q}) = \left(\begin{smallmatrix}
        \bm{X}_{\groupelementof{\text{F}}} \bm{X} \bm{X}_{\groupelementof{\text{F}}}^{-1} \\
        \bm{\rho}_{\configmanifold}(\groupelementof{\text{M}})\bm{q}
    \end{smallmatrix}\right),     
\end{equation}
where $\bm{X}\in\specialeucl{3}$ is the base pose represented as an element of $\specialeucl{3}$ and ${\bm{X}_{\groupelementof{\text{F}}} \in\specialeucl{3}}$ is the rigid-body transformation associated with the floating-base symmetry. The composed group action~\eqref{eq:morphosymm:base-config_action} therefore simultaneously transforms the base pose through conjugation in $\specialeucl{3}$ and permutes the joint coordinates according to the robot morphological symmetry.

\subsection{Rotational Symmetries}
\label{subsec:rotational_symmetries}
Many robotic manipulation tasks exhibit \emph{Euclidean isometries}, i.e., are equivariant under rigid-body transformations such as translations and rotations in task space.
Euclidean isometries preserve Euclidean distances between points in task space.
While translational equivariance is generally easy to implement, e.g., through standard convolutional architectures~\cite{zeng2021transporter}, rotational equivariance is typically more challenging to enforce.
Consequently, much of the literature~\cite{zeng2021transporter,wang2022:so2equivariant,huang2024leveraging} focuses specifically on rotational equivariance through the actions of the \emph{special orthogonal group} $\group_{\text{R}} \!=\! \specorthogonalgroup{n}$, $n\!\in\!\{2,3\}$ on the robot task space $\taskmanifold$. 
For a task space $\taskmanifold\subseteq\euclideanspace^n$, the corresponding group action is 
\begin{equation}
\groupactionon{\taskmanifold}{\text{R}}: \group_{\text{R}} \times \taskmanifold \rightarrow \taskmanifold, \quad\quad
    \groupactionon{\taskmanifold}{\text{R}}(\groupelementof{\text{R}}, \bm{x}) \!=\! 
        \bm{R}(\groupelementof{\text{R}})
        \bm{x},   
\end{equation}
where $\bm{R}(\groupelementof{\text{R}}) \!\in\! \specorthogonalgroup{n}$ is the rotation matrix associated with the element $\groupelementof{\text{R}}\in \group_{\text{R}}$. 
Since $\specorthogonalgroup{n}$ is a Lie group, its action admits an infinitesimal form through its Lie algebra $\specorthogonaliealgebra{n} = \{ \bm{\Omega} \!\in\! \mathbb{R}^{n \times n} | \bm{\Omega}^{T} \!=\! -\bm{\Omega} \}$, whose elements represent infinitesimal rotations. 
For every $\bm{\Omega} \in \specorthogonaliealgebra{n}$, the associated infinitesimal generator is the vector field $\infgenerator{\taskmanifold}{\bm{\Omega}}(\bm{x}) = \bm{\Omega} \bm{x}$.

\subsection{Task-induced Symmetries}

Task-induced symmetries~\cite{kallem2010task} describe transformations of $\bm{x} \in \taskmanifold$ that preserve a given cost function $c\!:\! \taskmanifold\! \to\! \mathbb{R}$.
Two states $\bm{x}_{1}, \bm{x}_{2} \!\in\! \taskmanifold$ are considered symmetric if $c(\bm{x}_{1}) \!=\! c(\bm{x}_{2})$.
Therefore, task-induced symmetries describe transformations within equivalence classes of states that leave the task cost $c$ unchanged.
Unlike morphological and rotational symmetries, task-induced symmetries are not specified a priori through a known group action.
Instead, they emerge implicitly from the structure of the task itself through the invariance of $c$.
Following~\cite{kallem2010task}, we focus on continuous task-induced symmetries. 
They are characterized locally by the infinitesimal directions in $\tasktangentspace{\bm{x}}$ along which $c$ remains constant, which form the basis of the space of motions in $\taskmanifold$ along which $c$ is invariant. Each basis component is a smooth vector field ${\vecfield{\taskmanifold}: \taskmanifold \to \tasktangentbundle}$ whose flow preserves $c$. 
A task-induced symmetry can be described by a group acting on $\taskmanifold$ only if the vector fields 
are complete (i.e., their flow exists for all times)~\cite[Thm. 20.16]{lee2012introduction} and closed under the Lie bracket~\cite[Ch.~8, p.~190]{lee2012introduction}. In this case, they define the infinitesimal generators $\infgenerator{\taskmanifold}{\bm{\xi}}$ of a Lie group $\group_{\text{T}}$ and their flows define the actions of $\group_{\text{T}}$ on $\taskmanifold$.
Specifically, the group action $\groupactionon{\taskmanifold}{\text{T}}\!:\! \group_{\text{T}} \!\times \!\taskmanifold \! \to \! \taskmanifold$ can be constructed numerically by integrating along $\infgenerator{\taskmanifold}{\bm{\xi}}$ as
\begin{equation}
    \label{eq:task_induced_group_action}
    \begin{split}
        & \groupactionon{\taskmanifold}{\text{T}}(g_{\text{T}}, \bm{x}) = \groupactionon{\taskmanifold}{\text{T}}(\operatorname{exp}(t\bm{\xi)}, \bm{x}) = \gamma(t, \bm{x}), \\
        \text{with}\quad & \frac{d \gamma}{dt} = \infgenerator{\taskmanifold}{\bm{\xi}}(\gamma(t, \bm{x})), ~~~ \gamma(0, \bm{x}) = \bm{x}.
    \end{split}
\end{equation}
By construction, the action~\eqref{eq:task_induced_group_action} preserves the cost $c$, i.e., $c(\groupactionon{\taskmanifold}{\text{T}}(g_{\text{T}}, \bm{x})) = c(\bm{x})$, $\forall g \in \group_{\text{T}}$, $\bm{x} \in \taskmanifold$.

\section{Cross-space Symmetry Transfer}
\label{sec:transferring_symmetries}

The families of symmetries introduced in Section~\ref{sec:catalog_of_symmetries} emerge jointly in various manipulation tasks, see, e.g., Fig.~\ref{fig:symmetries}. However, they act on different spaces with different group actions. 
In this section, we propose to transfer symmetries across configuration and task spaces in order to express them within a unified representation space, where they can subsequently be composed.
Transferring a symmetry requires preserving the effect of its action on the original space onto the target one.
Next, we describe the transfer of symmetries from $\configmanifold$ to $\taskmanifold$ and vice versa via the forward kinematics map ${f\!:\!\configmanifold\!\to\!\taskmanifold}$ for redundant robots, i.e., ${\text{dim}(\configmanifold)\!>\!\text{dim}(\taskmanifold)}$. 

\begin{figure}[t]
    \centering
    \includegraphics[width=\linewidth]{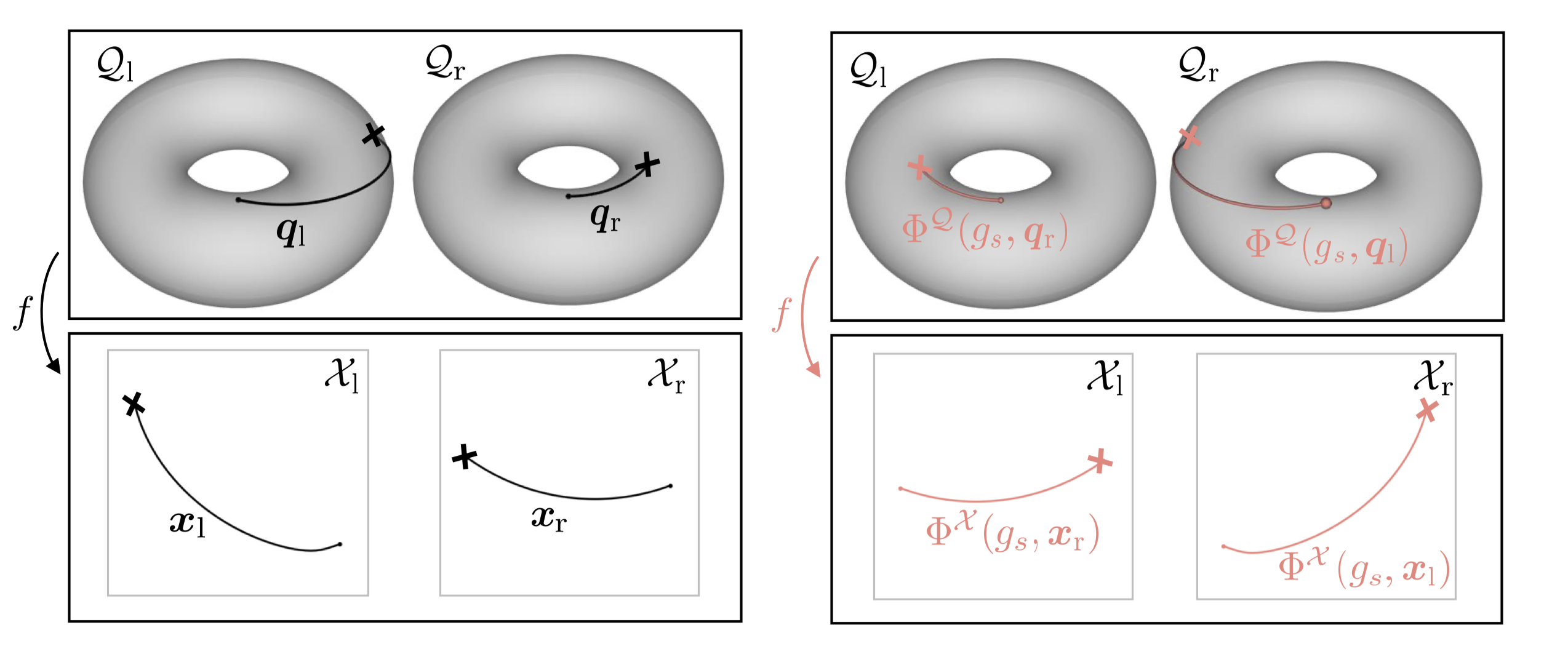}
    \caption{Descended symmetry. A symmetry in configuration space (\emph{top-left} to \emph{top-right}) induces a corresponding symmetry in task space (\emph{bottom-left} to \emph{bottom-right}) via the forward kinematics map.}
    \label{fig:descend}
    \vspace{-0.4cm}
\end{figure}

\subsection{Descending Symmetries}
\label{subsec:descending_symmetries}

We consider the action $\groupactiononblank{\configmanifold}\!:\! \group \! \times \configmanifold \! \rightarrow \! \configmanifold$ of a group $\group$ on the configuration space $\configmanifold$,  relating each configuration $\bm{q} \in \configmanifold$ to its symmetric counterparts $\bm{q}'\!=\!\groupactiononblank{\configmanifold}(g, \bm{q})$, $\forall g \in \group$. 
Here, we study the condition under which this symmetry induces a corresponding symmetry in task space, i.e., an equivalent relation between the associated task-space states $\bm{x}\!=\!f(\bm{q})$ and $\bm{x}'\!=\!f(\groupactiononblank{\configmanifold}(g, \bm{q})), \forall g \in \group$. 
When such a relation exists, we say that the symmetry \emph{descends} from configuration to task space, see Fig.~\ref{fig:descend}.
Formally, descending a symmetry requires constructing a group action $\groupactiononblank{\taskmanifold}\!:\! \group \! \times \taskmanifold \! \rightarrow \! \taskmanifold$ such that
\begin{equation}
    \label{eq:descend_equivariance_condition}
    \bm{x}'=f(\bm{q}^{\prime}) = 
    f(\groupactiononblank{\configmanifold}(g, \bm{q})) = 
    \groupactiononblank{\taskmanifold}(g, f(\bm{q})).
\end{equation}
holds $\forall\bm{q}\!\in\!\configmanifold$, $g\!\in\!\group$. 
In other words, a configuration-space symmetry descends to the task space if the forward kinematics map is $\group$-equivariant, i.e., fullfills the equivariance condition~\eqref{eq:equivariance_condition}.
Consequently, the task-space action $\groupactiononblank{\taskmanifold}$ is uniquely determined by $\groupactiononblank{\configmanifold}$ and $f$ through~\eqref{eq:descend_equivariance_condition} as
\begin{equation}
    \label{eq:descendend_task_space_group_action}
    \groupactiononblank{\taskmanifold}(g, \bm{x}) = 
    f(\groupactiononblank{\configmanifold}(g, \bm{q})).
\end{equation}
For example, the forward kinematics map $f$ is known to be $\group_{\text{M}}$-equivariant, with $\group_{\text{M}}$ denoting the morphological symmetry group~\cite{apraez2025morphological}. Therefore, morphological symmetries descend naturally and the descended group action is
\begin{equation}
    \label{eq:descended:morph_action}
    \groupaction{\text{M}}^{\taskmanifold}: \group_{\text{M}} \times \taskmanifold \rightarrow \taskmanifold, \quad\quad
    \groupaction{\text{M}}^{\taskmanifold}(\groupelementof{\text{M}}, \bm{x}) = \rho_{\taskmanifold}(\groupelementof{\text{M}}) \bm{x},
\end{equation}
where $\rho_{\taskmanifold}(\groupelementof{\text{M}})$ is the permutation of the end-effector poses.

\begin{mdframed}[hidealllines=true,backgroundcolor=lightlightgray,innerleftmargin=.1cm,innerrightmargin=.1cm,innertopmargin=.1cm,innerbottommargin=0.1cm,, roundcorner=2pt]
\begin{example}
    \label{example:descend_symmetry}
    We consider the dual-arm humanoid robot RB-Y1 (see Fig.~\ref{fig:symmetries}) with configuration space  $\configmanifold \!=\! \configmanifold_{l} \times \configmanifold_{r}$, where $\configmanifold_{l}, \configmanifold_{r}$ are the joint spaces of the $7$-DoF left and right arm, respectively. 
    The robot admits a morphological bilateral symmetry $\group_{\text{M}}$ described by the cyclic group ${\cyclicgroup{2} \!=\! \{e, g_{s}\}}$. The action of the non-trivial element $g_{s}$ on $\configmanifold$ permutes the joints $\bm{q}_{l} \in \configmanifold_{l}$ and $\bm{q}_{r} \in \configmanifold_{r}$ as
    \begin{equation}
            \groupactionon{\configmanifold}{\text{M}}(g_{s}, \bm{q}) = 
            \left(\begin{smallmatrix}
                \bm{0} & \bm{P} \\
                \bm{P} & \bm{0} \\
            \end{smallmatrix}\right)
            \left(\begin{smallmatrix}
                \bm{q}_{l} \\
                \bm{q}_{r} 
            \end{smallmatrix}\right) = 
            \left(\begin{smallmatrix}
                \bm{P}\bm{q}_{r} \\
                \bm{P}\bm{q}_{l} 
            \end{smallmatrix}\right),            
    \end{equation}
    where $\bm{P} = \operatorname{diag}(1, -1, -1, 1, -1, 1, -1)$ accounts for differences in the coordinate conventions of symmetric joints. 
    The descended group action induced through the forward kinematics onto the task space ${\taskmanifold \!=\! \euclideanspace^3 \!\times\! \euclideanspace^3}$ is
    \begin{equation}
            \groupaction{\text{M}}^{\taskmanifold}(g_{s}, \bm{x}) = 
            \left(\begin{smallmatrix}
                \bm{0} & \bm{P}_{\scriptscriptstyle\taskmanifold} \\
                \bm{P}_{\scriptscriptstyle\taskmanifold} & \bm{0}
            \end{smallmatrix}\right)
            \left(\begin{smallmatrix}
                \bm{x}_{l} \\
                \bm{x}_{r} \\
            \end{smallmatrix}\right) = 
            \left(\begin{smallmatrix}
                \bm{P}_{\scriptscriptstyle\taskmanifold}\bm{x}_{r} \\
                \bm{P}_{\scriptscriptstyle\taskmanifold}\bm{x}_{l}
            \end{smallmatrix}\right),
    \end{equation}
    with $\bm{P}_{\scriptscriptstyle\taskmanifold} \!=\! \operatorname{diag}(1, -1, 1)$ such that $f(\bm{P}\bm{q}_{l,r}) \!=\! \bm{P}_{\scriptscriptstyle\taskmanifold}\bm{x}_{l,r}$. Intuitively, the descended action reflects the positions of the two end-effectors across the robot sagittal plane, so that the task-space symmetry mirrors the bilateral symmetry of $\configmanifold$.\looseness-1
\end{example}
\end{mdframed}

\begin{figure}[t]
    \centering
        \includegraphics[width=\linewidth, trim=7.95cm 6.90cm 7.95cm 6.99cm, clip]{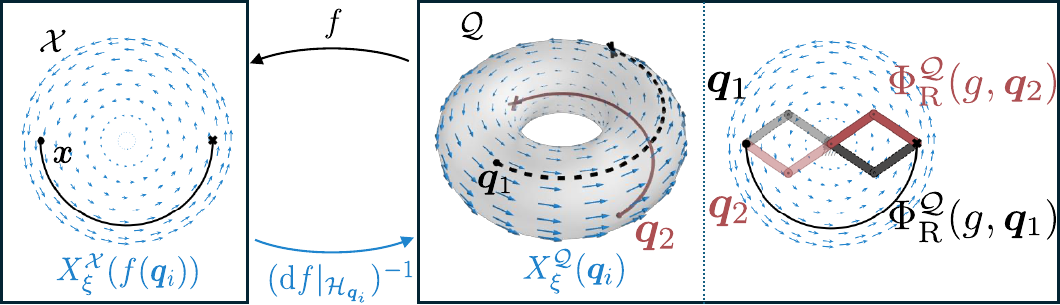}        
    \caption{Lifted symmetry. The infinitesimal generators of $\group$ in $\taskmanifold$ are horizontally lifted to $\configmanifold$, leading to different horizontally-lifted group actions for different $\bm{q}_{1},\bm{q}_2 \in$  $\configmanifold$ with $f(\bm{q}_{1})\!=\!f(\bm{q}_2)\!=\!\bm{x}$.
    }
    \label{fig:lifting_vf}
    \vspace{-0.4cm}
\end{figure}

\vspace{-0.2cm}
\subsection{Lifting Symmetries}
After characterizing descending symmetries, we proceed to study their \emph{lifting}, i.e., the transfer of task-space symmetries to $\configmanifold$. 
We consider a continuous symmetry described by the action $\groupactiononblank{\taskmanifold}\!:\! \group  \!\times\! \taskmanifold\! \rightarrow \!\taskmanifold$ of a Lie group $\group$ on the task space $\taskmanifold$. This action relates each task-space state $\bm{x}\in\taskmanifold$ to its symmetric counterparts $\bm{x}'\!=\!\groupactiononblank{\taskmanifold}(g, \bm{x})$, $\forall g \!\in\! \group$. 
Following the same reasoning as in Section~\ref{subsec:descending_symmetries}, lifting a symmetry would require constructing a group action ${\groupactiononblank{\configmanifold}\!:\! \group  \!\times\! \configmanifold\! \rightarrow \!\configmanifold}$ such that the inverse of the forward kinematics map is $\group$-equivariant.
However, for redundant robots, this problem is generally ill-posed due to the non-injectivity of the forward kinematics map, leading to the non-uniqueness of its inverse. 
As multiple joint configurations correspond to the same task-space state, a task-space symmetry does not uniquely determine a corresponding transformation in configuration space. Hence, directly constructing the lifted group action $\groupactiononblank{\configmanifold}$ through inverse kinematics becomes an under-determined problem. Next, we resolve this ambiguity by exploiting the additional geometric structure of the forward kinematics map.\looseness-1

\textbf{Forward kinematics and smooth submersion.}
In the following, we assume that: \textbf{(A1)} Our robots are redundant  and their configuration space $\configmanifold$ is a compact manifold, e.g., an $n$-dimensional torus; 
\textbf{(A2)} The configuration space is endowed with a Riemannian metric $\bm{M}(\bm{q})$;
\textbf{(A3)} The robot operates away from kinematic singularities, i.e., $\bm{J}({\bm{q}})$ has full row rank. Note these assumptions hold for a broad range of rigid-bodied robots and commonly-used task-space representations.
Under \textbf{(A1)}-\textbf{(A3)}, the forward kinematics map $f$ is a \emph{smooth submersion}~\cite{lee2012introduction}.
As explained next, this geometric structure allows us to lift task-space tangent vector fields to configuration-space tangent vector fields, which we subsequently leverage to lift symmetries.

\textbf{Horizontal lift.}
As a smooth submersion, the forward kinematics map decomposes each tangent space $\configtangentspace{\bm{q}}$ into vertical and horizontal subspaces (see Section~\ref{subsec:RiemannianBackground}).
The vertical subspace $\vertangent{\bm{q}} \!=\! \{\dot{\bm{q}} | \dot{\bm{x}}\!=\!\bm{J}(\bm{q})\dot{\bm{q}}\!=\!\bm{0}\}$ coincides with the nullspace of the Jacobian and contains the joint velocities that leave the end-effector state unchanged.
The horizontal subspace is ${\hortangent{\bm{q}} = (\mathcal{V}_{\bm{q}})^{\bot}}$. 
Since $f$ is a smooth submersion, the restricted differential
$
    \mathrm{d}f|_{\hortangent{\bm{q}}} \!:\! \hortangent{\bm{q}} \! \to \! \tasktangentspace{f(\bm{q})},
$
is a linear 
isomorphism, i.e., a structure-preserving invertible mapping.
Therefore, for each $\dot{\bm{x}} \in \tasktangentspace{f(\bm{q})}$, there is a unique ${\dot{\bm{q}}_{\hortangent{\bm{q}}} \in \hortangent{\bm{q}}}$, called the \emph{horizontal lift} of $\dot{\bm{x}}$ at $\bm{q}$, satisfying
\begin{equation}
    \label{eq:hor_lift_coordinatefree}
    \mathrm{d}f|_{\hortangent{\bm{q}}}(\dot{\bm{q}}_{\hortangent{\bm{q}}}) = \dot{\bm{x}}.
\end{equation}
In coordinates, the horizontal lift is given by
\begin{equation}
\label{eq:hor_lift_moore_penrose}
    \dot{\bm{q}}_{\hortangent{\bm{q}}} = \bm{J}(\bm{q})^{+}
    \dot{\bm{x}},
\end{equation}
where 
$\bm{J}^{+} \!=\! \bm{M}^{-1} \bm{J}^\trsp (\bm{J} \bm{M}^{-1} \bm{J}^\trsp)^{-1} $ is the generalized inverse of the Jacobian, where we dropped the dependency on $\bm{q}$ for readability.
Geometrically, the horizontal lift is the unique minimum-norm joint-space velocity, under the chosen Riemannian metric, that realizes the task-space velocity $\dot{\bm{x}}$.

\textbf{Lifting the vector fields.}
We take a vector-field perspective on task-space symmetries to lift them horizontally onto the configuration space. 
Using~\eqref{eq:infinitesimal_generator_from_group_action}, we characterize the group actions of continuous symmetries via their infinitesimal generators $\infgenerator{\taskmanifold}{\bm{\xi}}(\bm{x})$. 
At a configuration $\bm{q}$, the generator specifies a tangent vector $\infgenerator{\taskmanifold}{\bm{\xi}}(f(\bm{q})) \in \tasktangentspace{f(\bm{q})}$ describing the infinitesimal direction of the symmetric action.
Using~\eqref{eq:hor_lift_coordinatefree}-\eqref{eq:hor_lift_moore_penrose}, we obtain the unique horizontally-lifted vector field 
\begin{equation}
    \label{eq:inf_gen_hor_lift_general}
    \infgenerator{\configmanifold}{\bm{\xi}}(\bm{q}) = 
    (\mathrm{d}f|_{\hortangent{\bm{q}}})^{-1} 
    (\infgenerator{\taskmanifold}{\bm{\xi}}(f(\bm{q})))
    \quad \in \hortangent{\bm{q}},
\end{equation}
given, in coordinates, by
\begin{equation}
    \label{eq:inf_gen_hor_lift_moore_penrose}
    \infgenerator{\configmanifold}{\bm{\xi}}(\bm{q}) = 
    \bm{J}(\bm{q})^{+}
    \infgenerator{\taskmanifold}{\bm{\xi}}(f(\bm{q})).
\end{equation}
Importantly, we recover the task-space infinitesimal generator from the horizontally-lifted ones by construction, i.e.,
\begin{equation}
    \label{eq:_f-related_vec_fields}
    \mathrm{d}f (\infgenerator{\configmanifold}{\bm{\xi}}(\bm{q})) = 
    \infgenerator{\taskmanifold}{\bm{\xi}}(f(\bm{q})).
\end{equation}
so that $\infgenerator{\configmanifold}{\bm{\xi}}$ and $\infgenerator{\taskmanifold}{\bm{\xi}}$ are called $f$-related vector fields~\cite[Chap.~8]{lee2012introduction}. Fig.~\ref{fig:lifting_vf} illustrates the lifting of a vector field.

\textbf{From lifted vector fields to lifted infinitesimal generators.} 
To be an infinitesimal generator on $\configmanifold$, the lifted vector field~\eqref{eq:inf_gen_hor_lift_general} must preserve the composition rule of the Lie bracket on $\liealgebrablank$~\cite[Ch.~9,~Thm.~20.15]{lee2012introduction}.
Specifically, for any pair $\bm{\xi}, \bm{\zeta} \in \liealgebrablank$, the Lie bracket of their lifted vector fields $[\infgenerator{\configmanifold}{\bm{\xi}}, \infgenerator{\configmanifold}{\bm{\zeta}}]_{\configmanifold}$ must coincide with the horizontal vector field induced by their Lie bracket $[\bm{\xi}, \bm{\zeta}]_{\liealgebrablank}$, i.e.,
\begin{equation}
    \label{eq:lie_bracket_condition_configspace}
    [\infgenerator{\configmanifold}{\bm{\xi}}, \infgenerator{\configmanifold}{\bm{\zeta}}]_{\scriptscriptstyle \configmanifold}(\bm{q})
    =
    \infgenerator{\configmanifold}{[\bm{\xi}, \zeta]_{\scriptscriptstyle \liealgebrablank}}(\bm{q}).
\end{equation}
While $\infgenerator{\configmanifold}{[\bm{\xi}, \zeta]_{\scriptscriptstyle \liealgebrablank}}$ is horizontal by construction via~\eqref{eq:inf_gen_hor_lift_general}, the Lie bracket of two horizontal vector fields $\infgenerator{\configmanifold}{\bm{\xi}}$, $\infgenerator{\configmanifold}{\bm{\xi}}$ is not necessarily horizontal
~\cite[Sec. 34.3]{hassani2013mathematical}
In other words, for redundant robots, the Lie bracket $[\infgenerator{\configmanifold}{\bm{\xi}}, \infgenerator{\configmanifold}{\bm{\zeta}}]_{\scriptscriptstyle \configmanifold}(\bm q)$ may generate nullspace motions in $\mathcal{V}_{\bm q}$.
Therefore, the lifted vector fields are infinitesimal generators if the horizontal lift~\eqref{eq:hor_lift_moore_penrose} is chosen such that their Lie bracket is horizontal.

Such a horizontal lift always exists locally.
By the rank theorem~\cite[Thm.~4.12]{lee2012introduction}, around every $\bm{q} \!\in\! \configmanifold$, there exists a chart $(\mathcal{U}, \check{\bm{q}})$, $\mathcal{U} \subseteq \configmanifold$ with local coordinates $\check{\bm{q}}\!=\!(\bm{x}, \bm{r})$ such that $\check{f}(\bm{x}, \bm{r}) \!=\! \bm{x}$, i.e., motions along $\bm{r}$ are purely vertical. 
Choosing a block-diagonal metric $\bm{M}$ in $(\bm{x}, \bm{r})$, e.g., the identity, on $\configmanifold$ defines horizontal motions along $\bm{x}$ such that they are orthogonal to the vertical motions. Then, $[\infgenerator{\configmanifold}{\bm{\xi}}, \infgenerator{\configmanifold}{\bm{\zeta}}]_{\scriptscriptstyle \configmanifold}(\check{\bm{q}})$ is horizontal and~\eqref{eq:lie_bracket_condition_configspace} is locally satisfied. 
When considering joint constraints, the configuration manifold becomes an immersed submanifold $\tilde{Q}$ of $\configmanifold$~\cite[Def. 4.1]{bullo2005geometric}. The immersion $\iota\!:\! \tilde{\configmanifold} \!\to\! \configmanifold$ induces a unique $\iota$-related vector field $\infgenerator{\tilde{\configmanifold}}{\bm{\xi}}$ on $\tilde{\configmanifold}$, i.e.,  $\operatorname{d}\iota|_{\tilde{\bm{q}}}(\infgenerator{\tilde{\configmanifold}}{\bm{\xi}}(\tilde{\bm{q}}))\!=\! \infgenerator{\configmanifold}{\bm{\xi}}(\bm{q})$~\cite[Prop.~8.23]{lee2012introduction}.  If~\eqref{eq:lie_bracket_condition_configspace} holds, the naturality of Lie brackets~\cite[Prop.~8.30]{lee2012introduction} and the injectivity of $\iota$ imply $[\infgenerator{\tilde{\configmanifold}}{\bm{\xi}}, \infgenerator{\tilde{\configmanifold}}{\bm{\zeta}}]_{\scriptscriptstyle \tilde{\configmanifold}}(\tilde{\bm{q}})\!=\! \infgenerator{\tilde{\configmanifold}}{[\bm{\xi}, \zeta]_{\scriptscriptstyle \liealgebrablank}}(\tilde{\bm{q}})$.

\textbf{From lifted infinitesimal generators to lifted group actions.}
Given a chart $(\mathcal{U}, \check{\bm{q}})$, $\mathcal{U} \!\subseteq \!\configmanifold$ in which~\eqref{eq:lie_bracket_condition_configspace} holds, we aim to construct the lifted group action $\groupactiononblank{\configmanifold}$ from $\infgenerator{\configmanifold}{\bm{\xi}}$. 
While the horizontal lift defines a configuration-space infinitesimal generator, it does not guarantee the existence of a corresponding  group action.
This is achieved by ensuring that moving along the flow of $\infgenerator{\configmanifold}{\bm{\xi}}$ yields the same task-space motion as flowing along $\infgenerator{\taskmanifold}{\bm{\xi}}$. This requires \emph{(1)} the flow of $\infgenerator{\configmanifold}{\bm{\xi}}$ to be defined  $\forall\bm{q}\in\mathcal{U}$, and \emph{(2)} the flows of $\infgenerator{\configmanifold}{\bm{\xi}}$ and $\infgenerator{\taskmanifold}{\bm{\xi}}$ to preserve the equivariance~\eqref{eq:_f-related_vec_fields} of their generating vector fields. The former is always satisfied under (A1) since smooth vector fields on compact manifolds are complete~\cite[Cor.~9.17]{lee2012introduction}.
For the latter, the flows generated by $\infgenerator{\taskmanifold}{\bm{\xi}}$ and $\infgenerator{\configmanifold}{\bm{\xi}}$ are\looseness-1
\begin{align}
    & \tfrac{d}{dt}\gamma_{\bm{\xi}}^{\scriptscriptstyle\taskmanifold}(t, \bm{x}) = \infgenerator{\taskmanifold}{\bm{\xi}}(\gamma_{\bm{\xi}}^{\scriptscriptstyle\taskmanifold}(t, \bm{x})), \qquad \gamma_{\bm{\xi}}^{\scriptscriptstyle\taskmanifold}(0, \bm{x}) = \bm{x}, \\
    & \tfrac{d}{dt}\gamma_{\bm{\xi}}^{\scriptscriptstyle\configmanifold}(t, \bm{q}) = \infgenerator{\configmanifold}{\bm{\xi}}(\gamma_{\bm{\xi}}^{\scriptscriptstyle\configmanifold}(t, \bm{q})), \qquad~ \gamma_{\bm{\xi}}^{\scriptscriptstyle\configmanifold}(0, \bm{q}) = \bm{q},
    \label{eq:lifted_flow}
\end{align}
respectively. 
By the naturality of flows~\cite[Prop.~9.13]{lee2012introduction}, the equivariance of $f$-related vector fields extends to their flows,
\begin{equation}
    \label{eq:flows-f-relatedness}
    f(\gamma_{\bm{\xi}}^{\scriptscriptstyle\configmanifold}(t, \bm{q})) = 
    \gamma_{\bm{\xi}}^{\scriptscriptstyle\taskmanifold}(t, f(\bm{q})),
\end{equation}
and thus to their group actions $\groupactionon{\configmanifold}{\bm{\xi}}$ and $\groupactionon{\taskmanifold}{\bm{\xi}}$. This allows us to define the group action $\groupactiononblank{\configmanifold}(g, \bm{q}) = \gamma_{\bm{\xi}}^{\scriptscriptstyle\configmanifold}(t, \bm{q})$ as the unique horizontally-lifted group action of $\group$ on $\mathcal{U}$. If $\mathcal{U}=\configmanifold$, ~\eqref{eq:lie_bracket_condition_configspace} holds $\forall\bm{q} \in \configmanifold$ and $\groupactiononblank{\configmanifold}$ is a \emph{global} group action on $\configmanifold$.
Moreover, the horizontal lift is guaranteed to define a \emph{local} group action on the immersed submanifold $\tilde{\configmanifold}$ if $\tilde{\configmanifold}$ is compact, and a \emph{global} one if, in addition, $\mathcal{U}=\tilde{\configmanifold}$.
The periodicity of a periodic task-space group action with $\gamma_{\bm{\xi}}^{\taskmanifold}(T, \bm{x}) = \bm{x}$ is preserved , i.e., $\gamma_{\bm{\xi}}^{\configmanifold}(T, \bm{q}) = \bm{q}$, by~\eqref{eq:flows-f-relatedness} 
if $\bm{M}(\bm{q})$ is chosen such that $\mathcal{H}_{\bm{q}}$ is spanned by motions along $\bm{x}$, and thus the horizontally-lifted flow~\eqref{eq:lifted_flow} does not generate vertical motions.

\begin{mdframed}[hidealllines=true,backgroundcolor=lightlightgray,innerleftmargin=.1cm,innerrightmargin=.1cm,innertopmargin=.1cm,innerbottommargin=0.1cm,, roundcorner=2pt]
\begin{example}
    \label{example:lifting_symmetries}
    We consider an $n$-DoF planar robot whose task space $\taskmanifold = \mathbb{R}^2$ represents the end-effector position (see Fig.~\ref{fig:results}) and a rotational symmetry described by the group action
    $\groupaction{\text{R}}^{\taskmanifold}(g_{\text{R}}, \bm{x}) \!=\! 
    \bm{R}(g_{\text{R}})
    \bm{x}$ of $\group_{\text{R}} \!=\! \specorthogonalgroup{2}$ on $\taskmanifold$.
    From Section~\ref{subsec:rotational_symmetries}, the infinitesimal generators of $\specorthogonalgroup{2}$ are
    $
        \infgenerator{\taskmanifold}{\bm{\Omega}}(\bm{x}) = \bm{\Omega}\bm{x} =
        \left(\begin{smallmatrix}
            -\bm{x}_{2} \\
            \bm{x}_{1}
        \end{smallmatrix}\right) $.
    We obtain the configuration-space infinitesimal generator $\infgenerator{\configmanifold}{\bm{\Omega}}(\bm{q})$ from~\eqref{eq:inf_gen_hor_lift_moore_penrose} as
    \begin{equation}
        \infgenerator{\configmanifold}{\bm{\Omega}}(\bm{q}) = \bm{J}^{+}(\bm{q}) \infgenerator{\taskmanifold}{\bm{\Omega}}(f(\bm{q})).      
    \end{equation}
\end{example}
\end{mdframed}

\section{Symmetry Composition}
\label{sec:symmetry_composition}

In this section, we consider robot manipulation tasks exhibiting several symmetries, originating either in the configuration space $\configmanifold$ or the task space $\taskmanifold$. All symmetries are assumed to be expressed in a common space, either $\configmanifold$ or $\taskmanifold$. If not already the case, they are first transferred using Section~\ref{sec:transferring_symmetries}. Our goal is to learn a jointly-equivariant policy with respect to all symmetries at hand, i.e., the policy must satisfy the equivariance condition~\eqref{eq:equivariance_condition} simultaneously for the group actions $[\groupactionblank_1, \ldots, \groupactionblank_M]$. We study the compatibility conditions under which symmetries can be composed and characterize the resulting compositions based on their actions.

\subsection{Direct Product}
\label{subsec:direct_product}
Two groups $\group_{1}, \group_{2}$ can be composed into a product group ${\group \!= \!\group_{1} \!\times\! \group_{2} \!=\! \{g \!=\! (g_{1}, g_{2}) | g_{1} \in \group_{1}, g_{2} \in \group_{2} \}}$ via the direct product $\times$ if their actions $\groupaction{1}$, $\groupaction{2}$ commute, i.e.,
\begin{align}
    \label{eq:actions_commute}
    \groupaction{1}(g_{1}, \groupaction{2}(g_{2}, \bm{q})) = \groupaction{2}(g_{2}, \groupaction{1}(g_{1}, \bm{q})),
\end{align}
$\forall g_{1} \in \group_{1}, g_{2} \in \group_{2}, \bm{q} \in \manifold$.
Intuitively, the resulting transformation does not depend on the order in which the two symmetry actions are applied. For Lie groups, commutativity can be equivalently tested using their infinitesimal generators, i.e., the actions commute if
\begin{equation}
    \label{eq:lie_bracket_commutativity}
    [\infgenerator{\manifold}{\bm{\xi}}, \infgenerator{\manifold}{\bm{\zeta}}] = 0, ~~ \forall ~ \bm{\xi} \in \liealgebrablank_{1}, \bm{\zeta} \in \liealgebrablank_{2}.
\end{equation}
The composed group action $\groupactionblank$ is given as
\begin{equation}    
    \label{eq:direct_product_action}
    \groupactionblank(g, \bm{x}) = 
    \groupaction{1}(g_{1}, \groupaction{2}(g_{2}, \bm{x})) = 
    \groupaction{2}(g_{2}, \groupaction{1}(g_{1}, \bm{x})).
\end{equation}

\vspace{-0.2cm}
\begin{mdframed}[hidealllines=true,backgroundcolor=lightlightgray,innerleftmargin=.1cm,innerrightmargin=.1cm,innertopmargin=.1cm,innerbottommargin=0.1cm,, roundcorner=2pt]
\begin{example}
    \label{example:direct_product_composition}
    We consider the groups of planar rotations $\group_{1} \!=\! \specorthogonalgroup{2}$ and uniform scaling $\group_{2} \!=\! \mathbb{S}(2)$. Their group actions are $\groupaction{1}^{\taskmanifold} \!=\! \bm{R}(g_{1})\bm{x}$ and $\groupaction{2}^{\taskmanifold}\!=\!\lambda\bm{I}\bm{x}$ where $\lambda$ is a scaling factor. The corresponding infinitesimal generators are\looseness-1 
    $
        \infgenerator{\taskmanifold}{1}(\bm{x}) =
        \left(\begin{smallmatrix}
            -\bm{x}_{2} \\
            \bm{x}_{1}
        \end{smallmatrix}\right)$, and $
        \infgenerator{\taskmanifold}{2}(\bm{x}) =
        \left(\begin{smallmatrix}
            \bm{x}_{1} \\
            \bm{x}_{2}
        \end{smallmatrix}\right)$.
    The first generator induces infinitesimal rotations around the origin, while the second generates radial scaling motions. Their compatibility can be verified by computing their Lie bracket,
    \begin{equation}
        \begin{split}
        [\infgenerator{\taskmanifold}{1}, \infgenerator{\taskmanifold}{2}](\bm{x}) &= 
        \mathrm{d}\infgenerator{\taskmanifold}{1}|_{\bm{x}}\infgenerator{\taskmanifold}{2}(\bm{x}) - 
        \mathrm{d}\infgenerator{\taskmanifold}{2}|_{\bm{x}}\infgenerator{\taskmanifold}{1}(\bm{x}) \\
        & =
        \left(\begin{smallmatrix}
            -\bm{x}_{2} \\
            \bm{x}_{1}
        \end{smallmatrix}\right)
        -
        \left(\begin{smallmatrix}
            -\bm{x}_{2} \\
            \bm{x}_{1}
        \end{smallmatrix}\right) = 0.
        \end{split}
    \end{equation}
    Since the infinitesimal generators commute, the group actions commute and we compose the two groups through the direct product $\group \!=\! \specorthogonalgroup{2} \!\times\! \mathbb{S}(2)$ with combined group action
    \begin{equation}
        \groupactiononblank{\taskmanifold}(g, \bm{x}) = \bm{R}(g_{1})(\lambda\bm{I})\bm{x} = (\lambda\bm{I})\bm{R}(g_{1})\bm{x}.
    \end{equation}
\end{example}
\end{mdframed}

\subsection{Semi-Direct Product}
\label{subsec:semi_direct_product}
When two group actions do not commute, applying one action and then the other leads to a different transformation than applying them in the opposite order.
In this case, the two groups can still be combined if one acts on the other via automorphisms. This action is described by a homomorphism
$\varphi:\group_{1}\to\operatorname{Aut}(\group_{2})$,
where $\operatorname{Aut}(\group_{2})$ is the automorphism group of $\group_{2}$, i.e., the group of bijections from $\group_{2}$ to itself that preserve the group operation.
This implies that each element $g_{1}\!\in\!\group_{1}$ defines a structure-preserving transformation of $\group_{2}$.
Under this compatibility structure, the groups $\group_{1}$, $\group_{2}$ can be composed via the outer semi-direct product $\group \!=\! \group_{2} \rtimes \group_{1}$~\cite[Ch.~7]{lee2012introduction}, 
with elements $g \!=\! (g_{2}, g_{1}) \!\in\! \group$, yielding 
the composed group multiplication 
$(g_2, g_1) (g_2^{\prime}, g_1^{\prime}) \!=\! (g_2 \circ (\varphi(g_{1})(g_{2}^{\prime})), g_1 \circ g_1^{\prime}))$
and action
\begin{equation}
    \label{eq:semidirect_multlaw_and_action}
    \groupactionblank(g, \bm{q}) = 
    \groupaction{\group_{2}}(g_{2}, \groupaction{\group_{1}}(g_{1}, \bm{q})).  
\end{equation}
Unlike for the direct product, the first action is transformed by the second, reflecting their non-commutativity.
\begin{mdframed}[hidealllines=true,backgroundcolor=lightlightgray,innerleftmargin=.1cm,innerrightmargin=.1cm,innertopmargin=.1cm,innerbottommargin=0.1cm,, roundcorner=2pt]
\begin{example}
    \label{example:semi-direct_product_composition}
    We consider the reflectional $\group_{1} \!=\! \cyclicgroup{2}$ and rotational $\group_{2} \!=\! \specorthogonalgroup{2}$ symmetries acting on $\taskmanifold = \mathbb{R}^{2}$ via
    $\groupaction{\text{M}}^{\taskmanifold}(g_{1}, \bm{x}) \!=\! \bm{\rho}(g_{1}) \bm{x}$ and $\groupaction{\text{R}}^{\taskmanifold}(g_{2}, \bm{x}) \!=\! \bm{R}(g_{2}) \bm{x}$ with $\bm{\rho}(g_{1}) \!=\! 
        \left(\begin{smallmatrix}
            1 & 0 \\
            0 & -1
        \end{smallmatrix}\right)$
    and $\bm{R}(g_{2}) =
        \left(\begin{smallmatrix}
            \cos g_{2} & -\sin g_{2} \\
            \sin g_{2} & \cos g_{2}
        \end{smallmatrix}\right)$.
    From these actions, we observe that the non-trivial element of $\cyclicgroup{2}$ acts on $\specorthogonalgroup{2}$ through the automorphism
    $\varphi(g_{s})(\bm{R}(g_{2})) \!=\! \bm{\rho}(g_{s}) \bm{R}(g_{2}) \bm{\rho}(g_{s})^{-1} \!=\! \bm{R}(-g_{2})$.
    Geometrically, the reflection reverses the sense of rotation: A counterclockwise rotation in the original frame becomes a clockwise rotation in the reflected frame, and conversely.
    The two symmetry groups are composed via the semi-direct product as $\group \!=\! \specorthogonalgroup{2} \rtimes \cyclicgroup{2}$, which corresponds to the orthogonal group $\orthogonalgroup{2}$. From~\eqref{eq:semidirect_multlaw_and_action}, the composed action is\looseness-1
    \begin{equation}
        \groupactionblank^{\taskmanifold}((g_{2}, g_{1}), \bm{x}) 
        = 
        \bm{R}(g_{2})\bm{\rho}(g_{1})\bm{x}
        .
    \end{equation}
\end{example}
\end{mdframed}
\section{Experiments}
\label{sec:experimental_validation}

We experimentally validate the proposed cross-symmetry compositions on a letter-writing task with a simulated dual-arm planar robot, and on two real-world experiments on the dual-arm humanoid robot RB-Y1.
We endow the configuration manifolds $\configmanifold$ with the identity metric $\bm{M}(\bm{q}) \!=\! \bm{I}$ and learn joint-space policies via imitation learning, although our framework is applicable to other policy learning approaches.

\subsection{Letter-drawing Task with a Simulated Planar Robot}
\label{subsec:simulated-exps}
We consider a planar dual-arm robot with two $4$-DoF arms with configuration space $\configmanifold \!=\! \configmanifold_{\text{l}} \times \configmanifold_{\text{r}} \!=\! \mathbb{T}^4 \times \mathbb{T}^4$ and task space ${\taskmanifold \!=\! \taskmanifold_{\text{l}} \times \taskmanifold_{\text{r}} \!=\! \mathbb{R}^{2} \times \mathbb{R}^{2}}$. The robot is tasked to simultaneously trace the letters $\mathsf{C}$ and $\mathsf{N}$ with its left and right end-effectors.  
Our objective is to learn a policy capable of drawing these letters and symmetric variations thereof. We consider three symmetries: \emph{(1)} The bilateral \emph{morphological} symmetry of the arms, i.e., the cyclic-two group $\groupon{\configmanifold}_{\text{M}} \!=\! \cyclicgroup{2}$ acting on $\configmanifold$; \emph{(2)} A \emph{rotational} symmetry of $\groupon{\taskmanifold}_{\text{R}} \!=\! \specorthogonalgroup{2}$ acting on the letters in $\taskmanifold$; and \emph{(3)} A \emph{task-induced} scaling symmetry $\groupon{\taskmanifold}_{\text{S}} \!=\! \scalinggroup{2}$ acting on the letters in $\taskmanifold$.\looseness-1

We train a policy $\pi \!:\! \configmanifold \!\times\! \mathcal{S} \!\to\! \configtangentbundle$ that learns joint-space velocities $\dot{\bm{q}} \in \configtangentspace{\bm{q}}$ from configurations $\bm{q} \in \configmanifold$. We condition the network on a variable $\bm{s} \in \mathcal{S}$ that specifies the symmetric variant of the task to execute.
Each $\bm{s} = (\theta, \lambda, \sigma)$ is composed of a rotation angle $\theta\in[-\pi, \pi]$, a scaling factor $\lambda\in\mathbb{R}_{>0}$, and reflection index $\sigma\!\in\!\{1,-1\}$. 
In practice, the conditioning variable $\bm{s}$ can obtained from perception via a representation learning pipeline akin to~\cite{yang2024equivact}. We explicitly provide $\bm{s}$ to independently evaluate our cross-space symmetry composition framework.
The network is trained via imitation learning as in~\cite{perez2023stable} using $5$ demonstrations for training and $2$ for testing.

Following Sections~\ref{sec:transferring_symmetries} and~\ref{sec:symmetry_composition}, all symmetries are transferred to and composed in $\configmanifold$.
Then, we incorporate them into the policy via data augmentation.
The joint configurations and joint velocities of each demonstration are transformed via an action $\groupactiononblank{\configmanifold}(g, \bm{q})$ and its differential $\mathrm{d}\Phi^{\scriptscriptstyle \configmanifold}_{g}|_{\bm{q}}$, respectively.
We generate the augmented dataset using combinations of a finite subset of group elements per symmetry selected such that the corresponding transformations result in feasible robot states.
For the rotational symmetry, we discretize $\specialorth{2}$ using rotations $\bm{R}(\theta)$ with angular increment $\Delta \theta=\!30^{\circ}$ and $\Delta \theta\!=\!15^{\circ}$ for the training and testing trajectories, such that $\theta = k\Delta\theta \in [-\pi, \pi]$ with $k \in \mathbb{Z}$. Similarly, for $\scalinggroup{2}$, we use a predefined finite set of scaling factors $\lambda \in \{0.5, 0.6, 0.7, 0.8, 0.9, 1.0\}$. We use both elements $\{e,g_s\}\!\in\!\cyclicgroup{2}$ of the morphological symmetry group. The data augmentation resulted in $780$ training and $600$ testing trajectories.\looseness-1
\begin{table}[t]
    \centering
    \caption{RMSE of trained policies on test datasets augmented with an increasing number of different symmetric transformations.}
    \label{tab:mse}
    \setlength{\tabcolsep}{3pt}
    \resizebox{\linewidth}{!}{
    \renewcommand{\arraystretch}{1.3}
    \begin{tabular}{l cccc}
        \toprule
        & \multicolumn{4}{c}{\textit{Test datasets}} \\
        \cmidrule(lr){2-5}
        \textit{Policy}
            & Original
            & $\group_{\text{R}}$
            & $\group_{\text{RT}}$
            & $\group_{\text{MRT}}$ \\
        \midrule
        $\pi$
            & $0.203_{\pm 0.046}$
            & $3.841_{\pm 5.397}$
            & $3.008_{\pm 4.778}$
            & $2.804_{\pm 4.562}$ \\
        $\pi_{\group_\text{R}}$
            & $\bm{0.116_{\pm 0.006}}$
            & $\bm{0.134_{\pm 0.011}}$
            & $3.563_{\pm 9.660}$
            & $3.978_{\pm 10.921}$ \\
        $\pi_{\group_{\text{RT}}}$
            & $0.127_{\pm 0.012}$
            & $0.137_{\pm 0.013}$
            & $\bm{0.094_{\pm 0.016}}$
            & $3.635_{\pm 15.729}$ \\
        $\pi_{\group_{\text{MRT}}}$
            & $0.184_{\pm 0.034}$
            & $0.172_{\pm 0.041}$
            & $0.11_{\pm 0.020}$
            & $\bm{0.112_{\pm 0.018}}$ \\
        \bottomrule
    \end{tabular}
    }
    \vspace{-0.3cm}
\end{table}

\begin{figure}[t]
    \centering
    \begin{subfigure}[b]{0.24\linewidth}
        \includegraphics[
            width=\linewidth,
            trim={1.7cm 0.6cm 1.4cm 1.8cm},
            clip
            ]{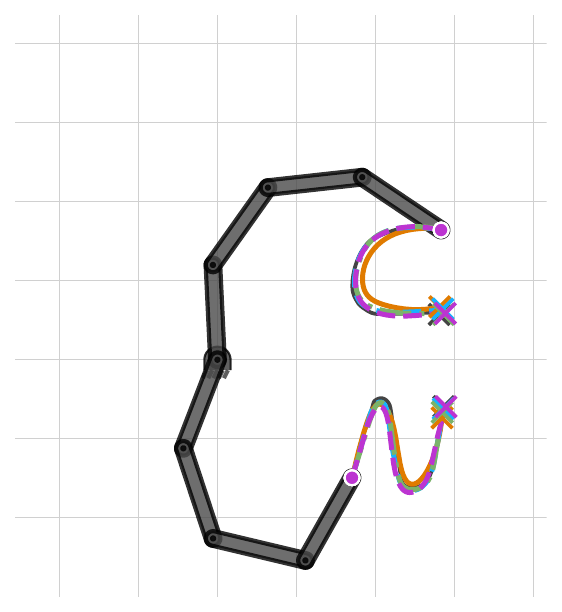}
        \caption{Original}
    \end{subfigure}
    \begin{subfigure}[b]{0.24\linewidth}
        \includegraphics[
            width=\linewidth,
            trim={1.7cm 0.6cm 1.4cm 1.8cm},            
            clip
            ]{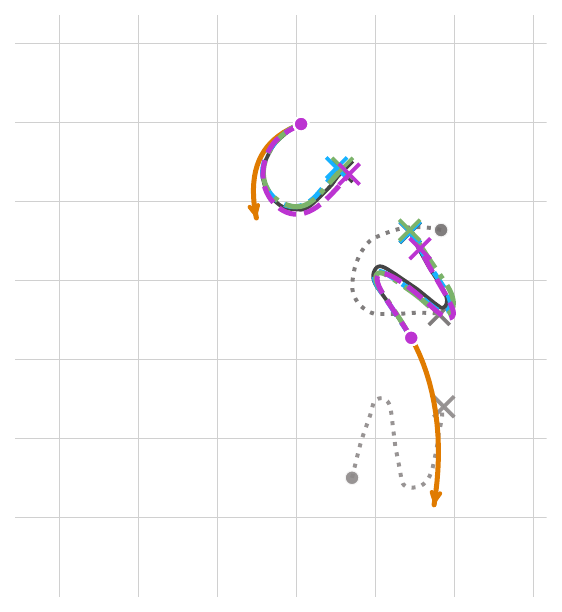}
        \caption{$\group_\text{R}$}
    \end{subfigure}
    \begin{subfigure}[b]{0.24\linewidth}
        \includegraphics[
            width=\linewidth,
            trim={1.7cm 0.6cm 1.4cm 1.8cm},            
            clip
            ]{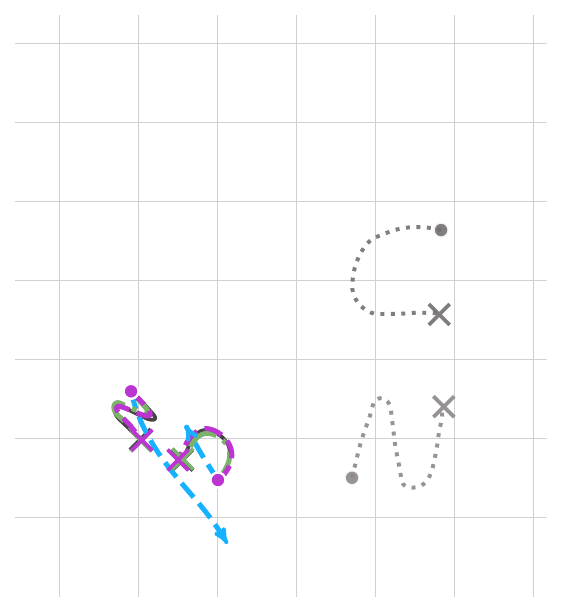}
        \caption{$\group_{\text{RT}}$}
    \end{subfigure}
    \begin{subfigure}[b]{0.24\linewidth}
        \includegraphics[
            width=\linewidth,
            trim={1.7cm 0.6cm 1.4cm 1.8cm},            
            clip            
            ]{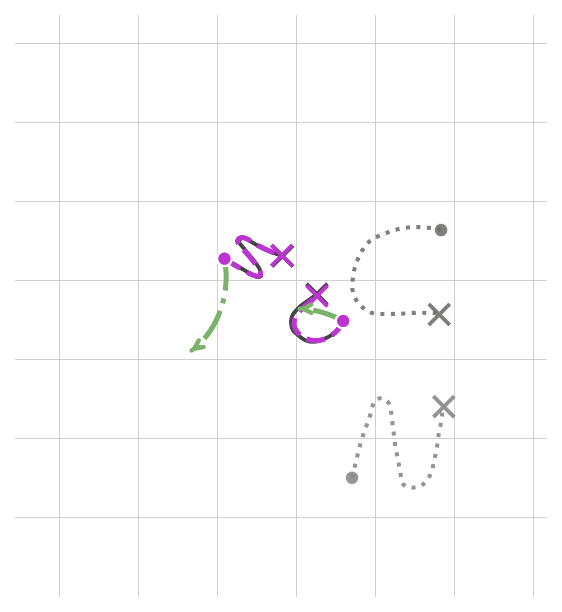}
        \caption{$\group_{\text{MRT}}$}
    \end{subfigure}
    \caption{Policies $\pi$ (\baselineline), $\pi_{\group_\text{R}}$ (\SOline), $\pi_{\group_{\text{RT}}}$ (\SOSline), $\pi_{\group_{\text{MRT}}}$ (\allsymline) evaluated on test trajectories (\blackline) obtained as symmetric transformations of the demonstrations (\grayline). The test trajectories are (a) the demonstrations, and (b)-(d) symmetric transformation thereof by different (composed) groups. Trajectories start and end are depicted as $\cdot$ and $\times$. Diverging trajectories are truncated and indicated by $\rightarrow$ and are excluded from the visualization of the next composition. }
    \label{fig:results}
    \vspace{-0.4cm}
\end{figure}

\textbf{Evaluation.} We evaluate the symmetry composition by considering $4$ policies: \emph{(1)} $\pi$, a base policy without symmetry augmentation; \emph{(2)} $\pi_{\group_{\text{R}}}$, encoding only rotational symmetry; \emph{(3)} $\pi_{\group_{\text{RT}}}$, encoding rotational and task-induced symmetries; and \emph{(4)} $\pi_{\group_{\text{MRT}}}$, encoding the full composition of morphological, rotational, and scaling symmetries. All policies are trained using the same optimization settings and procedures.

Fig.~\ref{fig:results} depicts trajectories generated by the learned policies under different symmetric transformations. As expected, $\pi$ only reproduces the original demonstrations and fails to generalize to symmetric variations. The rotationally-equivariant policy $\pi_{\text{R}}$ generalizes to rotated letters, but fails under scaling and morphological reflections. Incorporating the task-induced symmetry in $\pi_{\text{RT}}$ enables generalization to (jointly) rotated and scaled trajectories. Finally, the fully-symmetric policy $\pi_{\text{MRT}}$ generalizes consistently across all considered transformations, demonstrating the benefit of jointly encoding multiple symmetries.
Table~\ref{tab:mse} reports the mean and standard deviation across all tested trajectories (including diverging ones) of the  RMSE between the predicted and nominal trajectories starting from identical initial conditions, where nominal trajectories refer to both the original demonstrations and their symmetric transformations. We observe that incorporating the relevant symmetries consistently improves performance across all settings. In particular, whenever a policy lacks a symmetry present in the evaluation data, its prediction error increases significantly. The policy $\pi_{\group_{\text{MRT}}}$ performs well across all datasets, highlighting the effectiveness of cross-space symmetry compositions.


\begin{figure}[t]
    \centering
    \begin{subfigure}[b]{0.61\linewidth}
    \includegraphics[width=\textwidth]{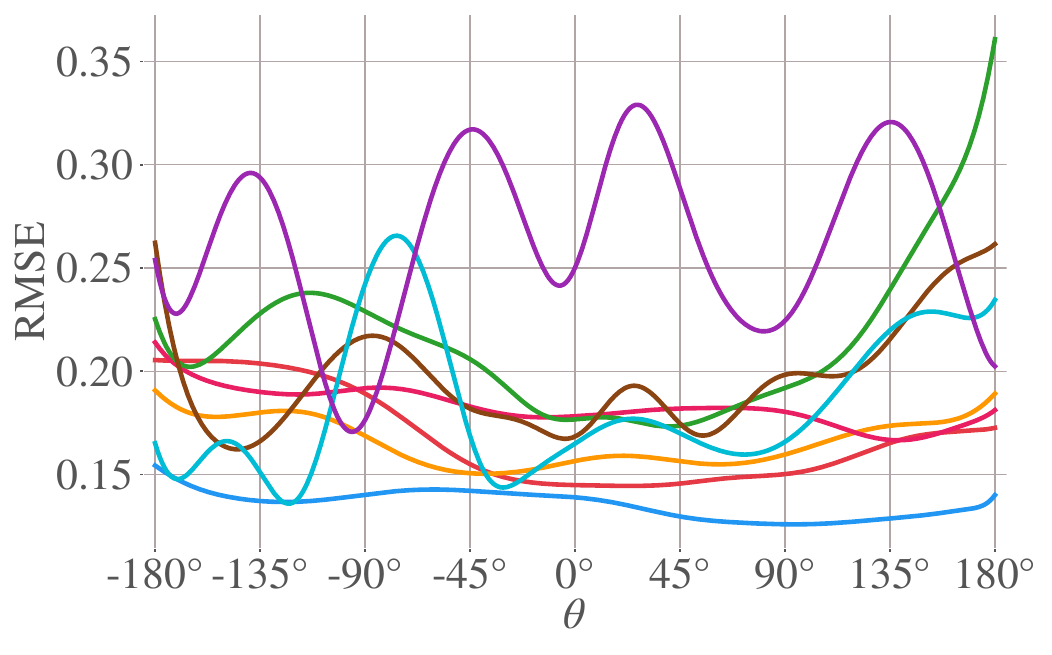}
    \vspace{-0.55cm}
    \caption{Task-space RMSE.}
    \label{fig:augm_density_quantitative}
    \end{subfigure}
    \begin{subfigure}[b]{0.37\linewidth}
        \includegraphics[width=\linewidth]{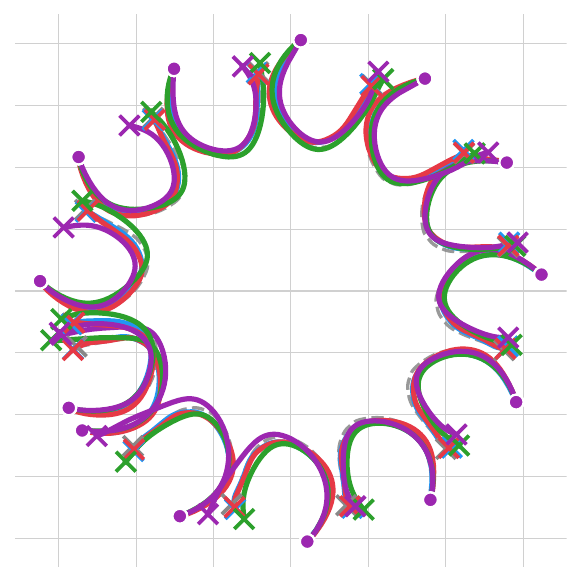}
        \caption{Test trajectories.}
        \label{fig:augm_density_qualitative}
    \end{subfigure}
    \caption{Evaluation of policies $\pi_{\group_{\text{R}}}$ trained with varying augmentation intervals. (a) RMSE for policies trained with intervals $5^{\circ}$ (\SOFiveDeg), $10^{\circ}$ (\SOTenDeg), $15^{\circ}$ (\SOFifteenDeg), $30^{\circ}$ (\SOThirtyDeg), $45^{\circ}$ (\SOFortyFiveDeg), $60^{\circ}$ (\SOSixtyDeg), $75^{\circ}$ (\SOSeventyFiveDeg), $90^{\circ}$ (\SONinetyDeg) and evaluated on $\theta$-rotated letters.
    (b) Trajectories for policies $\pi_{\group_\text{R}}$ trained with augmentation intervals of $5^{\circ}$ (\SOFiveDeg), $10^{\circ}$ (\SOTenDeg), $45^{\circ}$ (\SOFortyFiveDeg), $90^{\circ}$ (\SONinetyDeg) evaluated on test trajectories (\blackline) at $30^{\circ}$ intervals. Trajectories start and end are depicted as $\cdot$ and $\times$.
    }
    \vspace{-0.2cm}
\end{figure}

\textbf{Effect of data augmentation resolution on policy generalization.}  
We train several policies $\pi_{\group_{\text{R}}}$ in $\configmanifold$ using different discretization steps for the $\specialorth{2}$ group actions in $\taskmanifold$. 
Fig.~\ref{fig:augm_density_quantitative} reports the task-space RMSE between predicted and nominal trajectories as a function of the angle $\theta$.
As expected, denser augmentation consistently improves performance across all rotations. In contrast, policies trained with coarser augmentation exhibit larger performance variations, with errors increasing as the queried rotation moves farther away from the nearest augmented training sample.
Fig.~\ref{fig:augm_density_qualitative} shows that, even under relatively coarse augmentation in $\taskmanifold$, the learned policies still produce trajectories in $\configmanifold$ that mostly preserve the intended letter shapes across the whole range of rotations.

\begin{figure}[t]
    \centering
    \begin{subfigure}[b]{0.315\linewidth}
        \includegraphics[
            width=\linewidth,
            height=0.6\linewidth,
            trim={25.0cm 0pt 20pt 10.0cm},
            clip
        ]{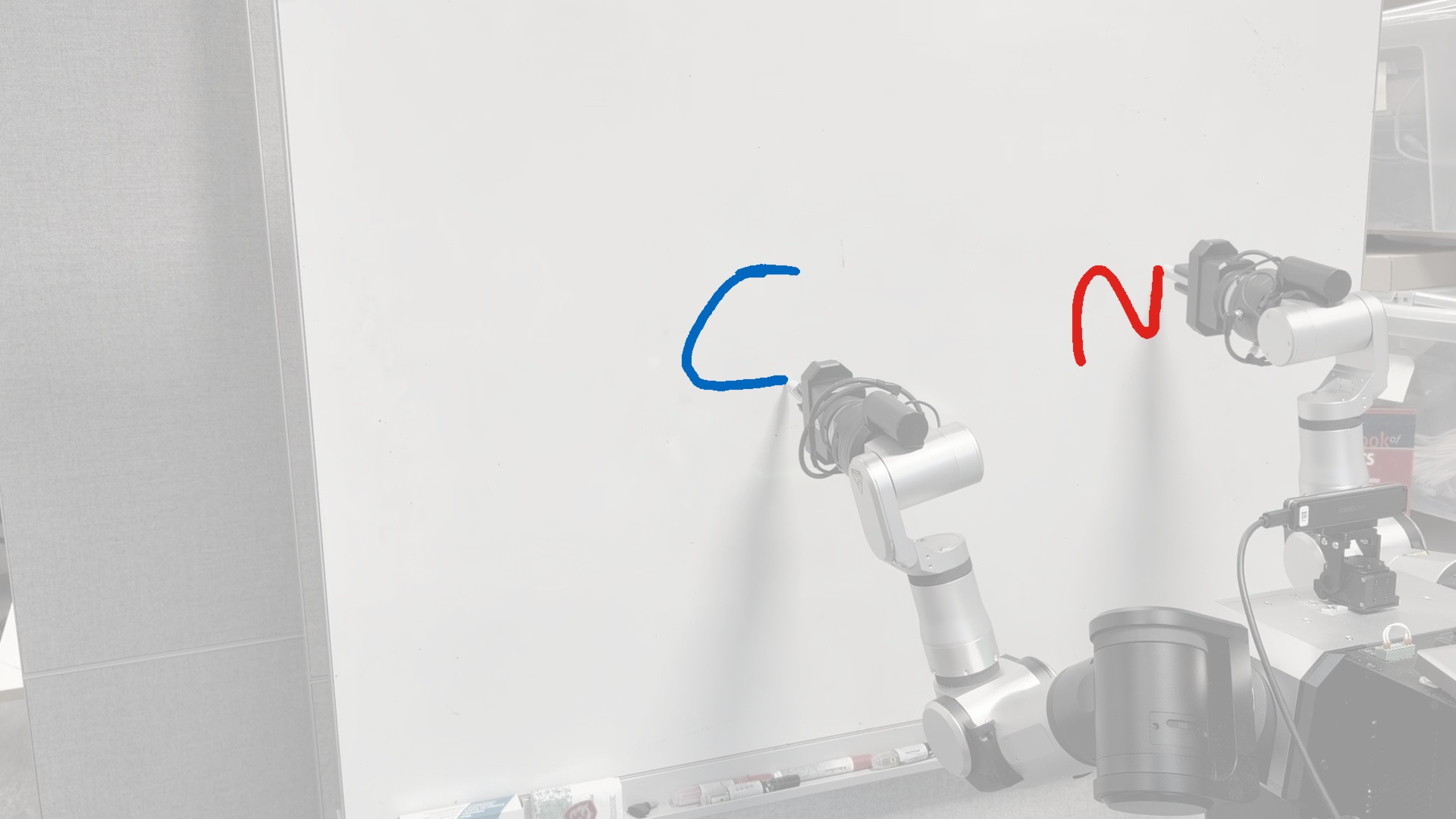}
        \caption{ $\bm{s} = (0, 1, 1)$}
    \end{subfigure}
    \begin{subfigure}[b]{0.315\linewidth}
        \includegraphics[
            width=\linewidth,
            height=0.6\linewidth,
            trim={25.0cm 0pt 20pt 10.0cm},
            clip
        ]{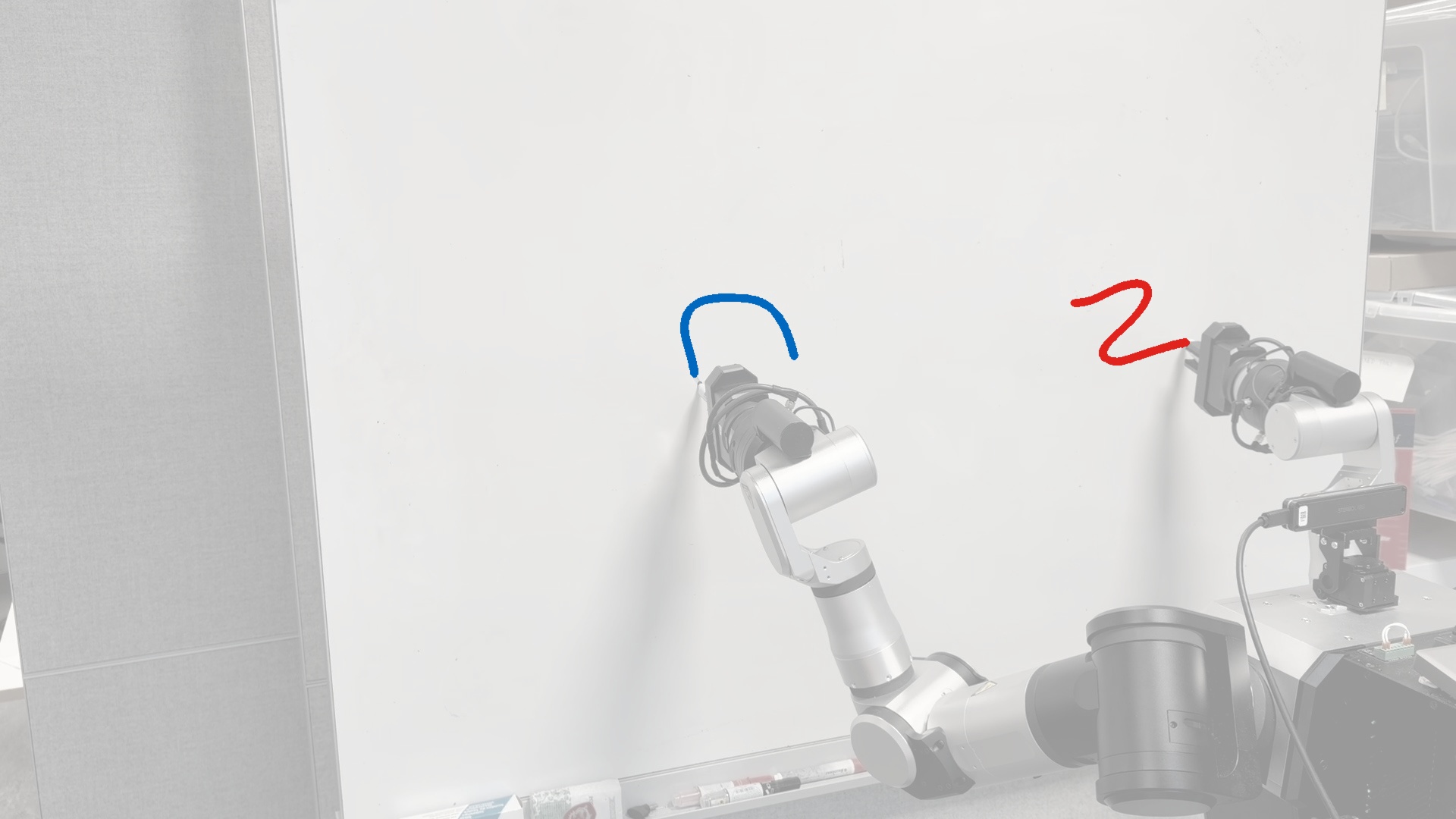}
        \caption{ $\bm{s} = (90^{\circ}, 1, 1)$}
    \end{subfigure}
    \begin{subfigure}[b]{0.315\linewidth}
        \includegraphics[
            width=\linewidth,
            height=0.6\linewidth,
            trim={25.0cm 5.0cm 20pt 18.0cm},
            clip
            ]{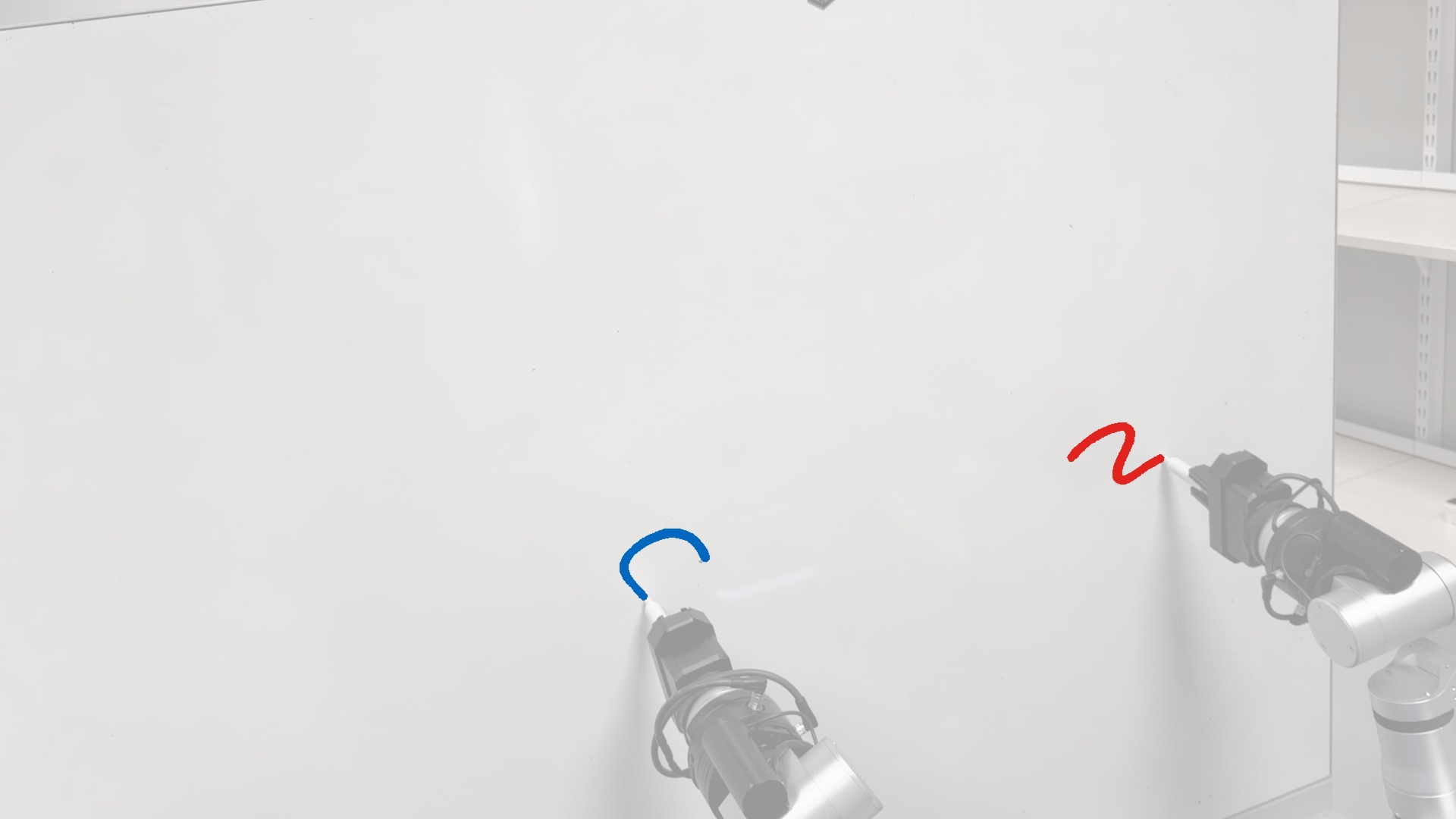}
        \caption{ $\bm{s} = (90^{\circ}, 0.5, 1)$}
    \end{subfigure}
    \\
    \begin{subfigure}[b]{0.315\linewidth}
        \includegraphics[
            width=\linewidth,
            height=0.6\linewidth,
            trim={5.0cm 0.0cm 10.0cm 10.0cm},
            clip            
            ]{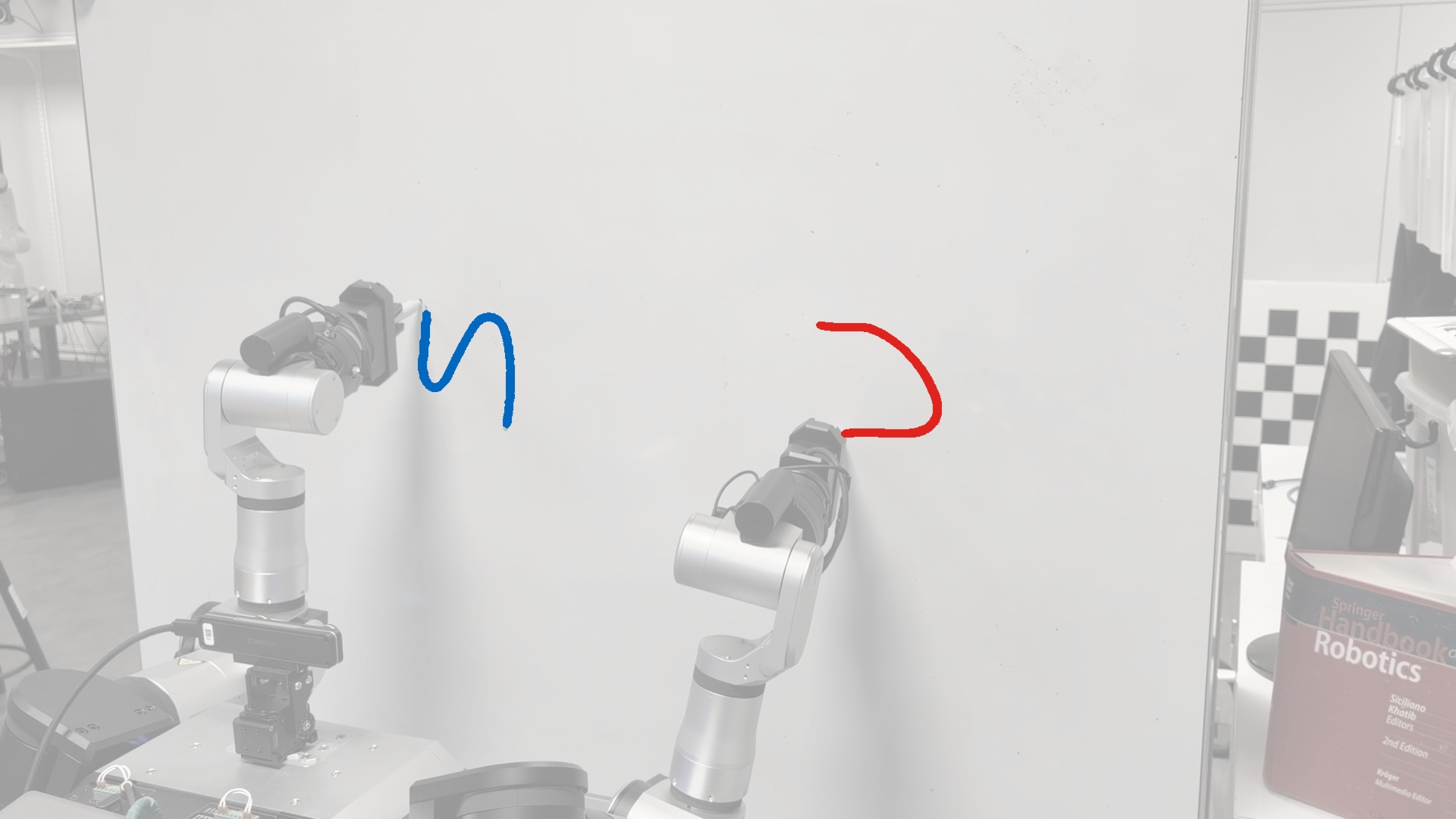}
        \caption{ $\bm{s} = (0, 1, -1)$}
    \end{subfigure}
    \begin{subfigure}[b]{0.315\linewidth}
        \includegraphics[
            width=\linewidth,
            height=0.6\linewidth,
            trim={15.0cm 0.0cm 0.0cm 10.0cm},
            clip            
            ]{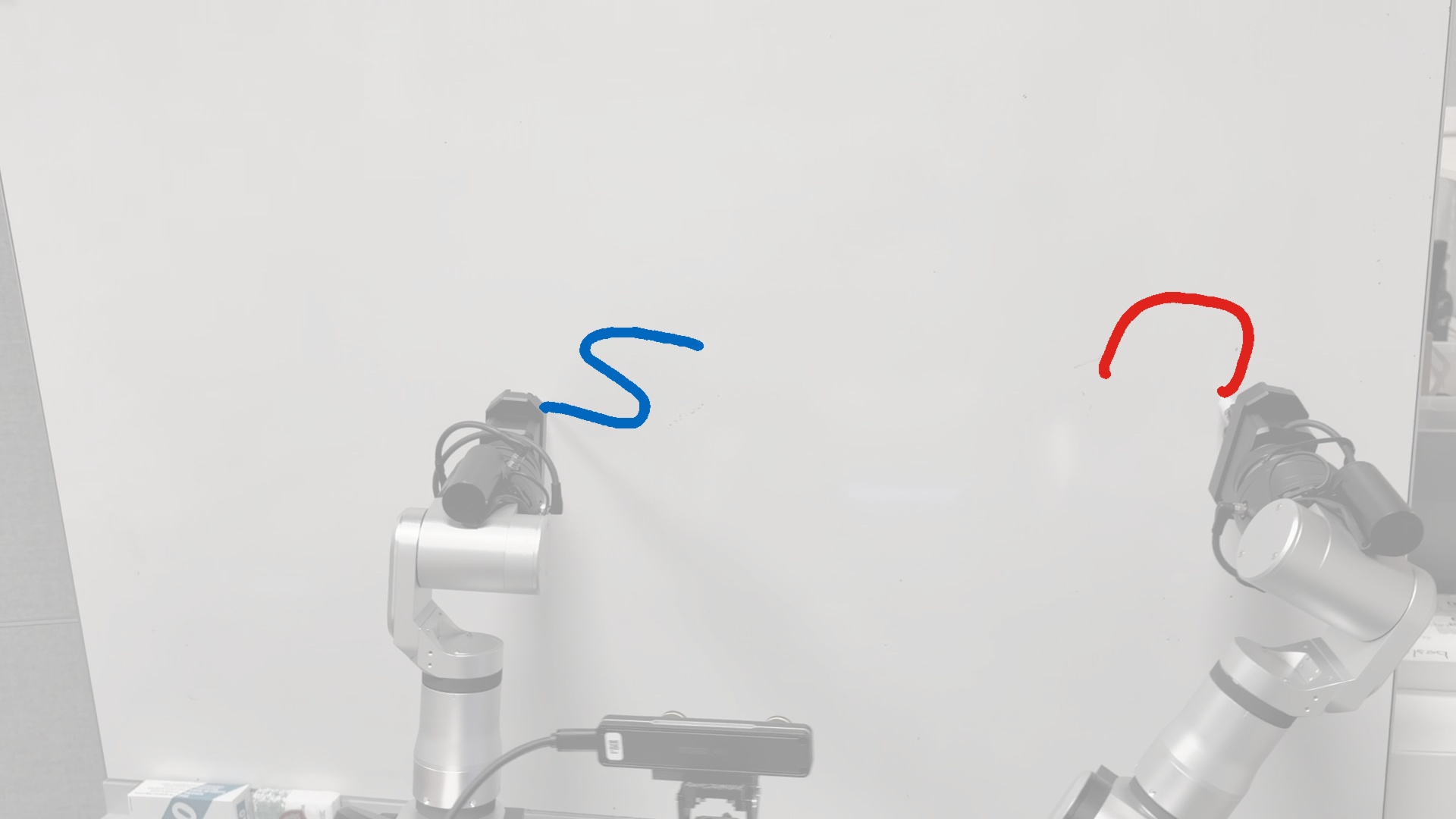}
        \caption{$\bm{s} = (90^{\circ}, 1, -1)$}
    \end{subfigure}
    \begin{subfigure}[b]{0.315\linewidth}
        \includegraphics[
            width=\linewidth,
            height=0.6\linewidth,
            trim={15.0cm 0.0cm 0.0cm 8.0cm},
            clip            
            ]{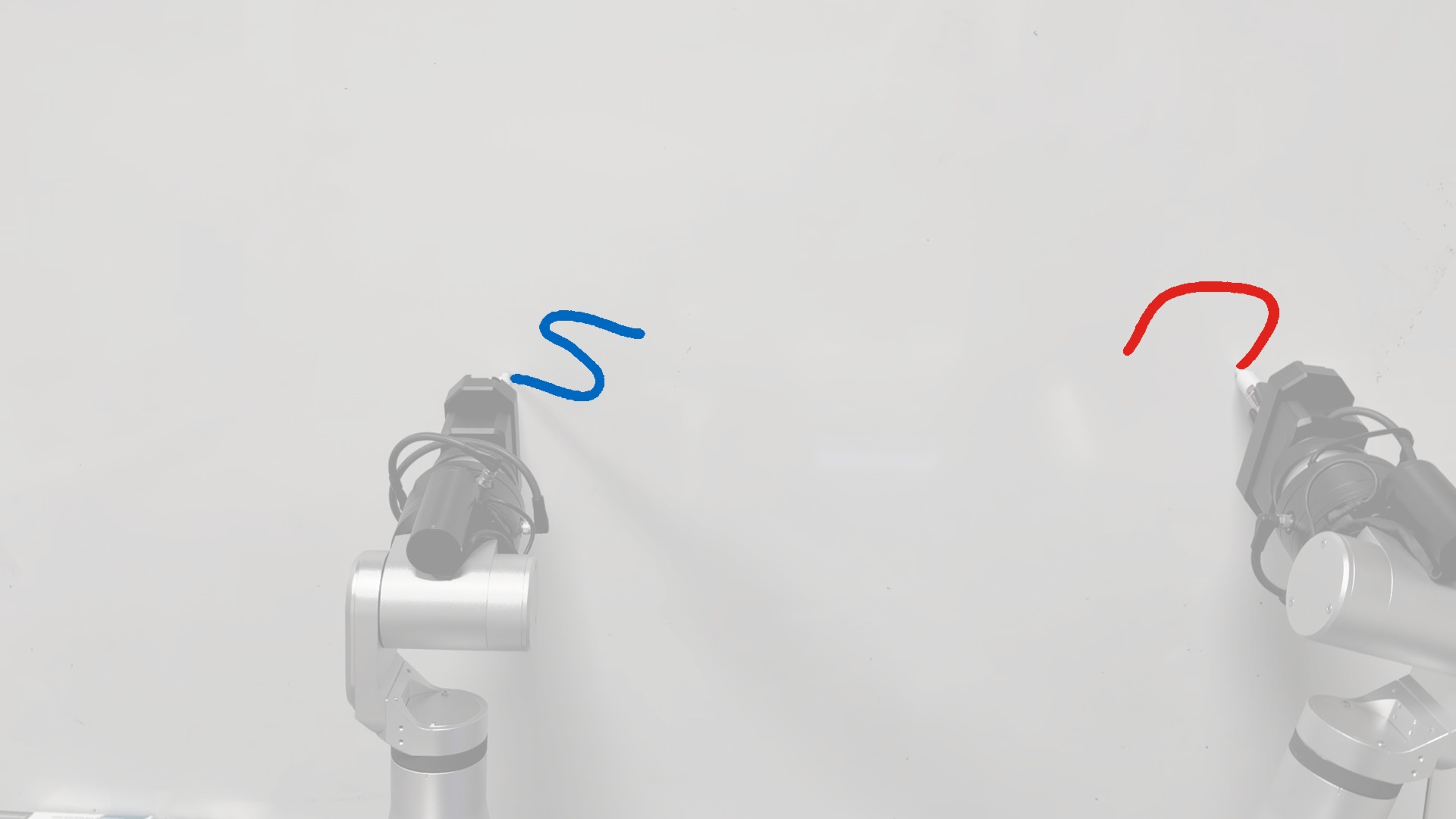}
        \caption{ $\bm{s} = (90^{\circ}, 0.5, -1)$}
    \end{subfigure}    
    \caption{Real-world letter-drawing experiment. The robot draws rotated and scaled letters using its left (\SOFiveDeg) and right (\SOTenDeg) arms.}
    \label{fig:results:rby1-letters}
    \vspace{-0.4cm}
\end{figure}

\begin{figure}[t]
    \centering
    \begin{subfigure}[b]{0.315\linewidth}
        \includegraphics[
            width=\linewidth,
            trim={2.0cm 0.0cm 5.0cm 0.0cm},
            clip            
            ]{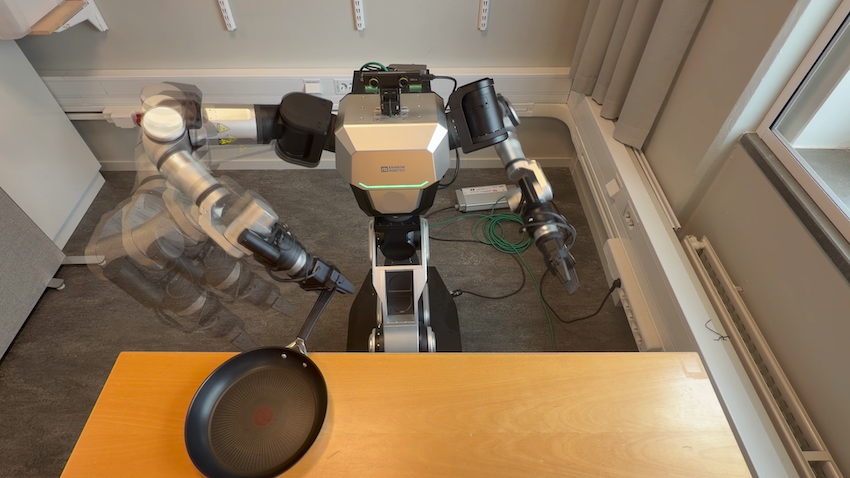}
        \caption{ $\bm{s} = (30^{\circ}, 1, 1)$}
    \end{subfigure}
    \begin{subfigure}[b]{0.315\linewidth}
        \includegraphics[
            width=\linewidth,
            trim={2.0cm 0.0cm 5.0cm 0.0cm},
            clip            
            ]{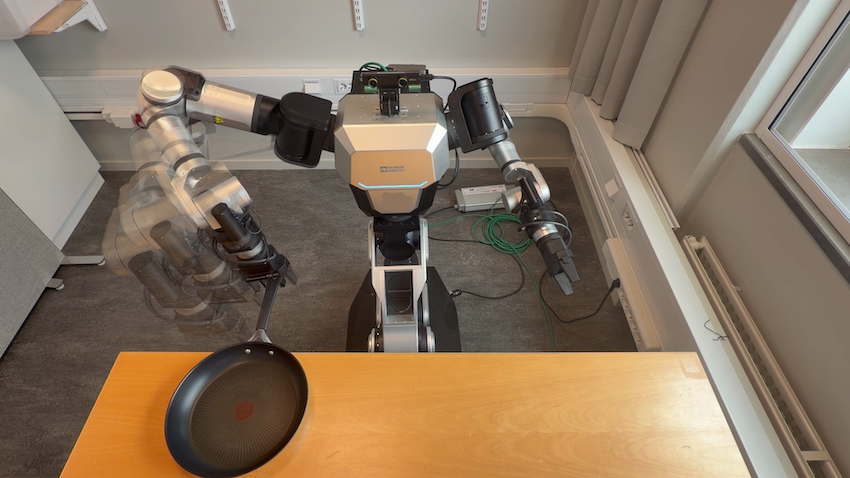}
        \caption{ $\bm{s} = (0^{\circ}, 1, 1)$}
    \end{subfigure}
    \begin{subfigure}[b]{0.315\linewidth}
        \includegraphics[
            width=\linewidth,
            trim={2.0cm 0.0cm 5.0cm 0.0cm},
            clip            
            ]{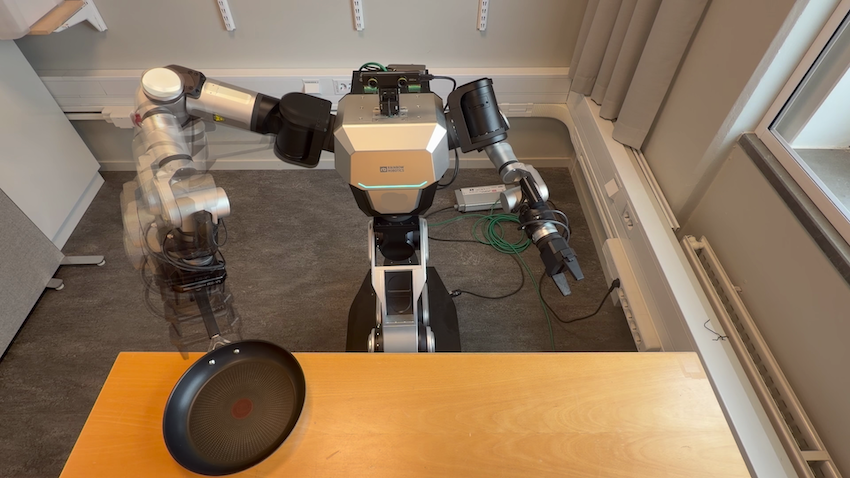}
        \caption{ $\bm{s} = (-30^{\circ}, 1, 1)$}
    \end{subfigure}
    \\
    \begin{subfigure}[b]{0.315\linewidth}
        \includegraphics[
            width=\linewidth,
            trim={2.0cm 0.0cm 5.0cm 0.0cm},
            clip            
            ]{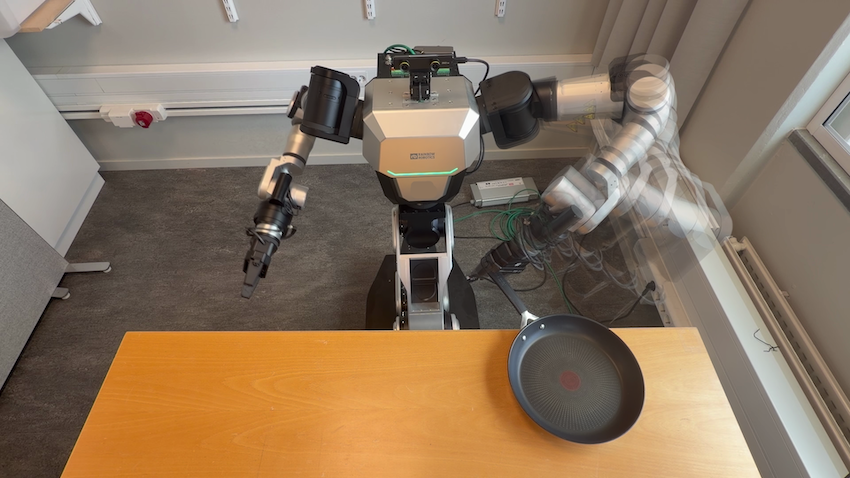}
        \caption{ $\bm{s} = (-30^{\circ}, 1, -1)$}
    \end{subfigure}
    \begin{subfigure}[b]{0.315\linewidth}
        \includegraphics[
            width=\linewidth,
            trim={2.0cm 0.0cm 5.0cm 0.0cm},
            clip            
            ]{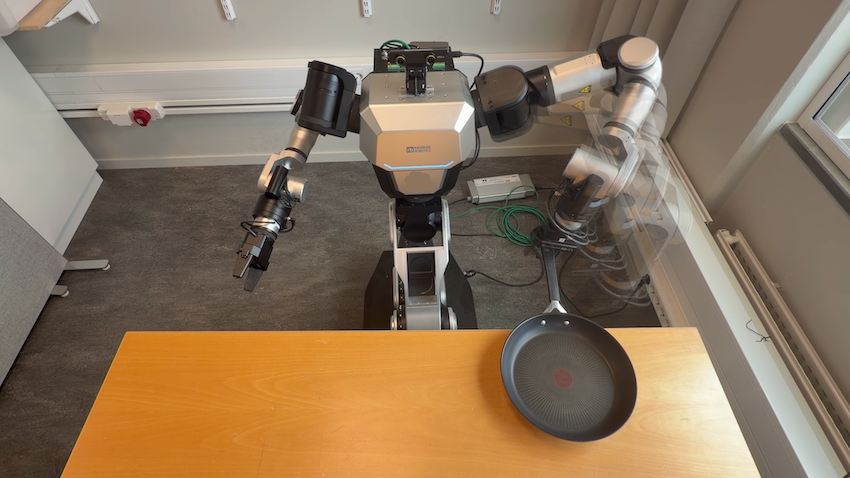}
        \caption{$\bm{s} = (0^{\circ}, 1, -1)$}
    \end{subfigure}
    \begin{subfigure}[b]{0.315\linewidth}
        \includegraphics[
            width=\linewidth,
            trim={2.0cm 0.0cm 5.0cm 0.0cm},
            clip            
            ]{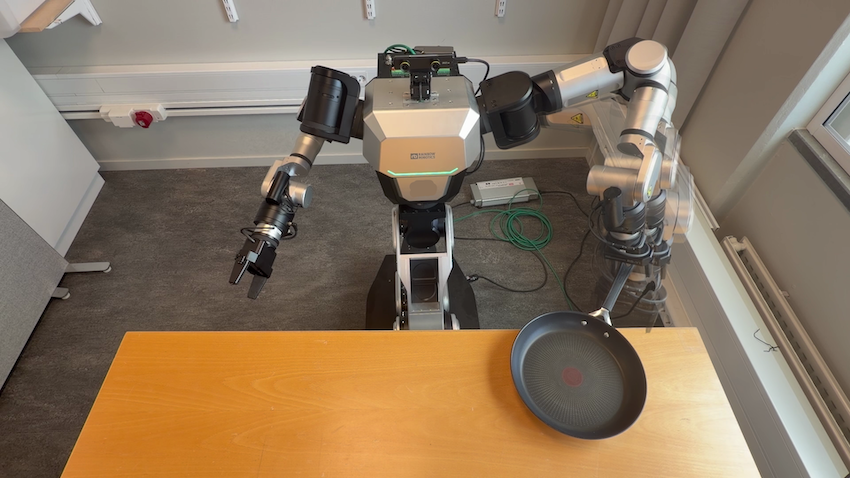}
        \caption{\scriptsize $\bm{s} = (30^{\circ}, 1, -1)$}
    \end{subfigure}    
    \caption{Real-world pan-grasping experiment. The robot grasps the rotated pan with either its left (\emph{bottom}) or right (\emph{top}) arm.}
    \label{fig:results:rby1-pan}
    \vspace{-0.6cm}
\end{figure}

\subsection{Real-Robot Experiments}

We validate the proposed cross-space symmetry compositions through two real-world experiments on the dual-arm humanoid robot RB-Y1.
Similar to Section~\ref{subsec:simulated-exps}, we train joint-space symmetry-conditioned policies $\pi \!:\! \configmanifold \!\times\! \mathcal{S} \! \to \! \configtangentbundle$.

\textbf{Letter-drawing task.} The first experiment extends the dual-arm letter-drawing task to the real robot. In this setting, the policy is trained on demonstrations augmented with the full composed symmetry group $\group_{\text{MRT}}$, enabling simultaneous rotational, scaling, and morphological equivariance across the two $7$-DoF arms of the RB-Y1.
Fig.~\ref{fig:results:rby1-letters} shows executions of the learned policy generated under different symmetry-conditioning values $\bm{s}$.
The learned policy successfully generalized across the $17$ considered symmetry transformations, with a task-space trajectory RMSE of $0.033_{\pm 0.007}$, allowing the robot to trace rotated and scaled letters using either arm. 

\textbf{Pan-grasping task.} For the second experiment, the robot must grasp the handle of a rotated pan using either arm.
We collected $2$ demonstrations with the right arm at  pan orientation of $\theta\!=\!0$.
To enable generalization across pan rotations and arms interchangeability, the dataset is augmented using rotational symmetries via the group $\groupon{\taskmanifold}_{\text{R}} \!=\! \specialorth{2}$ acting on $\taskmanifold$ and the bilateral morphological symmetries using the group $\groupon{\configmanifold}_{\textbf{M}} \!=\! \cyclicgroup{2}$ acting on $\configmanifold$.
As illustrated in Fig.~\ref{fig:results:rby1-pan}, the resulting equivariant policy successfully generalizes to $5$ unseen pan orientations and executes grasping motions with both arms with a task-space trajectory RMSE of $0.017_{\pm 0.004}$.

\section{Conclusion}
\label{sec:conclusion}

This paper introduces \emph{cross-space symmetry compositions}, a framework for learning robot policies that are jointly equivariant to multiple symmetries arising across configuration and task spaces. Building on the differential-geometric structure of the forward kinematics map, we transfer symmetries across spaces by \emph{descending} configuration-space symmetries to task space and \emph{lifting} task-space symmetries to configuration space. Moreover, we characterize the compatibility conditions under which transferred symmetries can be systematically composed through direct and semi-direct products.
Our framework unifies several important symmetry families commonly encountered in robotics, including morphological, rotational, and task-induced symmetries, within a common mathematical formalism. Our experiments demonstrated that jointly leveraging multiple symmetries substantially improves policy generalization across task variations.\looseness-1

While validated with joint-space policies learned via imitation learning, our framework can also be used for other learning paradigms, e.g., reinforcement learning, and that the descending direction of the transfer allows learning equivariant task-space policies. As future work, we will learn equivariant visuomotor policies with cross-space symmetry compositions, thereby integrating our framework with equivariant perception modules operating on point clouds~\cite{yang2024equivact} via the descending operation. 
We will also investigate the integration of cross-space symmetry compositions into equivariant neural architectures, to enforce equivariance at the architectural level rather than weakly through data augmentation.\looseness-1

\bibliographystyle{IEEEtran}
\bibliography{settings/references.bib}

\end{document}